\documentclass[journal]{IEEEtran}
\usepackage{times}
\usepackage{epsfig}
\usepackage{graphicx}
\usepackage{amsmath}
\usepackage{amssymb}
\usepackage{amsthm}

\usepackage{comment}
\usepackage{enumitem}
\usepackage{pifont}
\usepackage{url}
\usepackage{booktabs}
\usepackage{makecell}
\usepackage{multirow}
\usepackage{caption}
\usepackage{gensymb}

\usepackage[linesnumbered,lined,boxed,commentsnumbered,ruled]{algorithm2e}
\usepackage[colorlinks,linkcolor=black,anchorcolor=black,citecolor=black,urlcolor=black]{hyperref}
\usepackage{algorithmic}
\usepackage[caption=false]{subfig}
\renewcommand{\paragraph}[1]{\noindent\textbf{#1}~~}

\usepackage{bbding}
\usepackage{tabularx}
\usepackage{rotating}

\ifCLASSOPTIONcompsoc
  \usepackage[nocompress]{cite}
\else
  \usepackage{cite}
\fi

\ifCLASSINFOpdf
\else
\fi

\begin{document}
%
\title{Context-Aware Mixup for Domain Adaptive Semantic Segmentation}

\author{Qianyu~Zhou, 
       Zhengyang~Feng,
       Qiqi~Gu, 
       Jiangmiao~Pang, \\
       Guangliang~Cheng, 
       Xuequan~Lu$^{\dagger}$, 
       Jianping~Shi, 
       Lizhuang~Ma$^{\dagger}$
\thanks{Manuscript received 14 April, 2022. revised 07 July 2022. accepted
04 September 2022. Date of publication xx xx 2022; date of current version
xx xx 2022. This work is supported by National Key Research and Development Program of China (2019YFC1521104), National Natural Science Foundation of China (72192821, 61972157), Shanghai Municipal Science and Technology Major Project  (2021SHZDZX0102), Shanghai Science and Technology Commission (21511101200, 22YF1420300), and Art major project of National Social Science Fund (I8ZD22).}
\thanks{Q. Zhou, F. Zheng, Q. Gu and L. Ma are with the Department of Computer Science and Engineering, Shanghai Jiao Tong University, Shanghai 200240, China (e-mail: \{zhouqianyu, zyfeng97, miemie\}@sjtu.edu.cn and ma-lz@cs.sjtu.edu.cn.)}

\thanks{J. Pang is with Shanghai AI Laboratory, China (e-mail: pangjiangmiao@gmail.com).}

\thanks{G. Cheng and J. Shi are with SenseTime Research, Beijing, China (e-mail:guangliangcheng2014@gmail.com, shijianping@sensetime.com).}

\thanks{X. Lu is with the School of Information Technology, Deakin University,
Victoria 3216, Australia (e-mail: xuequan.lu@deakin.edu.au).}

\thanks{
$^\dagger$ Joint corresponding author.}

}

\markboth{IEEE TRANSACTIONS ON CIRCUITS AND SYSTEMS FOR VIDEO TECHNOLOGY,~Vol.~X, No.~X, X}%
{Shell \MakeLowercase{\textit{et al.}}: Bare Demo of IEEEtran.cls for IEEE Journals}

\IEEEpubid{\begin{minipage}{\textwidth}\ \\ \\ \\ \\[8pt] \centering
Copyright $\copyright$ 2022 IEEE. Personal use of this material is permitted. However, permission to use this material for any other purposes must be obtained from the IEEE by sending an email to pubs-permissions@ieee.org.
\end{minipage}}

\maketitle

\begin{abstract}
Unsupervised domain adaptation (UDA) aims to adapt a model of the labeled source domain to an unlabeled target domain.  Existing UDA-based semantic segmentation approaches always reduce the domain shifts in pixel level, feature level, and output level. However, almost all of them largely neglect the contextual dependency, which is generally shared across different domains, leading to less-desired performance. In this paper, we propose a novel Context-Aware Mixup (CAMix) framework for domain adaptive semantic segmentation, which exploits this important clue of context-dependency as explicit prior knowledge in a fully end-to-end trainable manner for enhancing the adaptability toward the target domain.  Firstly, we present a contextual mask generation strategy by leveraging the accumulated spatial distributions and prior contextual relationships. The generated contextual mask is critical in this work and will guide the context-aware domain mixup on three different levels. Besides, provided the context knowledge, we introduce a significance-reweighted consistency loss to penalize the inconsistency between the mixed student prediction and the mixed teacher prediction, which alleviates the negative transfer of the adaptation, \emph{e.g.,} early performance degradation.
Extensive experiments and analysis demonstrate the effectiveness of our method against the state-of-the-art approaches on widely-used UDA benchmarks. 
\end{abstract}


%
\IEEEpeerreviewmaketitle

\section{Introduction}
\IEEEPARstart{S}{emantic} segmentation aims to assign a semantic label to each pixel for a given image. 
Over the past few years, researchers have made great efforts to explore a variety of CNN methods~\cite{chen2018deeplab,long2015fully,chen2018encoder,zhao2017pspnet,ji2020encoder,zhou2016computation,zhao2022novel,Tan_2022_TPAMI_mirror} trained on a large-scale segmentation dataset~\cite{Pascal,COCO,cordts2016cityscapes,Tan_2021_TIP_NightCity} to tackle this problem. However, building such a large annotated dataset is both cost-expensive and time-consuming due to the process of annotating pixel-wise labels~\cite{cordts2016cityscapes}. A natural idea to overcome this bottleneck is using synthetic data~\cite{stephan2016gtav,ros2016synthia} to supervise the segmentation model instead of real data. However, the existing domain gap between the synthetic images~\cite{stephan2016gtav,ros2016synthia} and real images~\cite{cordts2016cityscapes} often leads to a significant performance drop when the learned source models are directly applied to the unlabelled target data. 

\begin{figure}[t]
\centering
\includegraphics[scale=1.1]{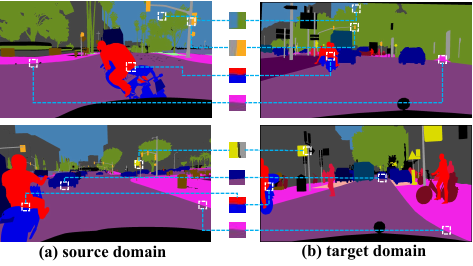}\caption{ Previous domain adaptive semantic segmentation methods largely neglect the shared context-dependency across different domains when adapting from the source domain to the target domain, and could result in less-desired performance and severe negative transfer. We observe that exploiting contexts as explicit prior knowledge is essential for enhancing the adaptability toward the target domain during the adaptation.}
\label{fig1}
\vspace{-4mm}
\end{figure}

To address this issue, various unsupervised domain adaptation (UDA) techniques for semantic segmentation have been proposed to reduce the domain gap in pixel level~\cite{Cycada,FDA,BDL,LTIR,CrDoCo,guo2021label,PCEDA,LDR}, feature level~\cite{SIBAN,CBST,CRST,STAR,FADA,DADA,liu2021bapa} and output level~\cite{AdaptSegNet,CLAN,SIM,CLANv2,IntraDA,APODA}.  Among them, the most common practices are based on adversarial learning~\cite{AdaptSegNet,CLAN,SIM,CLANv2,IntraDA}, self-training~\cite{CBST,CRST,BDL,TPLD,IAST,LSE}, consistency regularization~\cite{choi2019self,SEANet,tranheden2020dacs,liu2021bapa}, and entropy minimization~\cite{ADVENT,MaxSquare,FDA}. These works mainly focused on utilizing common knowledge, \emph{e.g.,} appearances, scales, textures, weather, \emph{etc.,} to narrow down the domain gap.

Nevertheless, context-dependency across different domains has been very sparsely exploited so far in UDA, and how to transfer such cross-domain context still remains under-explored.
As shown in Fig.~\ref{fig1}, we observe that the source and target images usually share similar semantic contexts, \emph{e.g.,} a rider is over the bicycle or motorcycle, the sidewalk is beside the road, and such context knowledge is crucial, particularly when adapting from the source domain to the target domain.  The lack of context will lead to less-desired performance during the adaptation and even severe negative transfer, \emph{e.g.,} early performance degradation during the adaptation process.
Previous works \cite{tranheden2020dacs,french2019semi,french2020milking} neglect the context dependency in the domain mixup, and we observe that images synthesized by these methods often violate the contextual relationships between objects. For instance, only un-occluded parts of mid-range objects are copied onto the irrelevant classes of other images. Imagine how strange it is
to see a pedestrian standing on top of a car or to see the sky through a hole in a building. 
Thus, the lack of such context information results in category confusion and label contamination in the mixed results (\emph{e.g.,} Fig.~\ref{fig:visulization_mixup}).
Besides, the state-of-the-art approaches of domain adaptive semantic segmentation heavily depend on adversarial learning, image-to-image translation, or self-training, and most of them need to fine-tune or re-train the models in many offline stages, which are quite complex and hard to converge and cannot be trained in an end-to-end manner.  

Motivated by the above facts, we propose a novel perspective of domain adaptive semantic segmentation that identifies context-dependency across domains as explicit prior domain knowledge when adapting from the source domain to the target domain. As such, we present a context-aware domain mixup (\textbf{CAMix}) framework to explicitly explore and transfer cross-domain contexts for domain adaptation. Our whole framework is fully end-to-end trainable and easy to implement.

\begin{figure}[t]
\centering
\vspace{-7mm}
\includegraphics[scale=0.55]{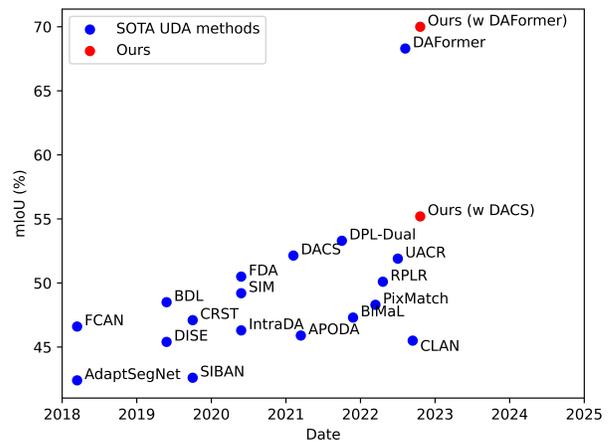}
\caption{
Progress of UDA for semantic segmentation on GTAV~\cite{stephan2016gtav} $\rightarrow$ Cityscapes~\cite{cordts2016cityscapes}. Our CAMix shows superior performance compared to the state-of-the-art methods.
}
\label{fig:uda_progess}
\end{figure}

The proposed CAMix framework consists of two key components: contextual mask generation (CMG) and significance-reweighted consistency loss (SRC). 
Specifically, CMG firstly generates a contextual mask by selectively leveraging the accumulated spatial distribution of the source domain and the contextual relationship of the target domain. This mask is critical in our work and will act as prior knowledge to guide the context-aware domain mixup on three different levels, \emph{i.e.,} input level, output level, and significance mask level. {Notice that the significance mask is a mask that we define to indicate where the pixels are credible.} This contextual mask respectively mixes the input images, the labels, and the corresponding significance-masks to narrow down the domain gap. In addition, we introduce an SRC loss on the significance mask level to alleviate the negative transfer, \emph{e.g.,} early performance degradation, during the adaptation process. In particular, we calculate a significance mask with the help of the target predictive entropy and its dynamic threshold. Then, we mix the target and the source significance masks using the context knowledge and utilize the mixed significance mask to reweigh the consistency loss. Extensive experiments with analysis demonstrate that CAMix achieves superior performance against the state-of-the-art methods, as shown in Fig.~\ref{fig:uda_progess}.

Our contributions are summarized as follows. 

$\bullet$  From a new perspective, we propose a novel \textit{context-aware mixup} (CAMix) framework for domain adaptive semantic sefmentation, which exploits context dependency across domains as explicit prior domain knowledge for further improving the adaptability towards the target domain. 

$\bullet$  We present a \textit{contextual mask generation} strategy, which leverages the spatial distribution of the source domain and the contextual relationship of the target domain for guiding the context-aware domain mixup on three different levels. Besides, we introduce a \textit{significance-reweighted consistency loss}, which alleviates the adverse impacts of the adaptation procedure, \emph{e.g.,} early performance degradation, under the guidance of context.
    
$\bullet$ Extensive experiments with analysis demonstrate the effectiveness of our method on two challenging UDA benchmarks. Our CAMix can be easily plugged into existing UDA frameworks,  \emph{e.g.,} DACS~\cite{tranheden2020dacs} and DAFormer~\cite{hoyer2022daformer}, and achieve consistent improvements over the state-of-the-art methods.

\begin{figure*}[t]
\centering
\includegraphics[scale=0.18]{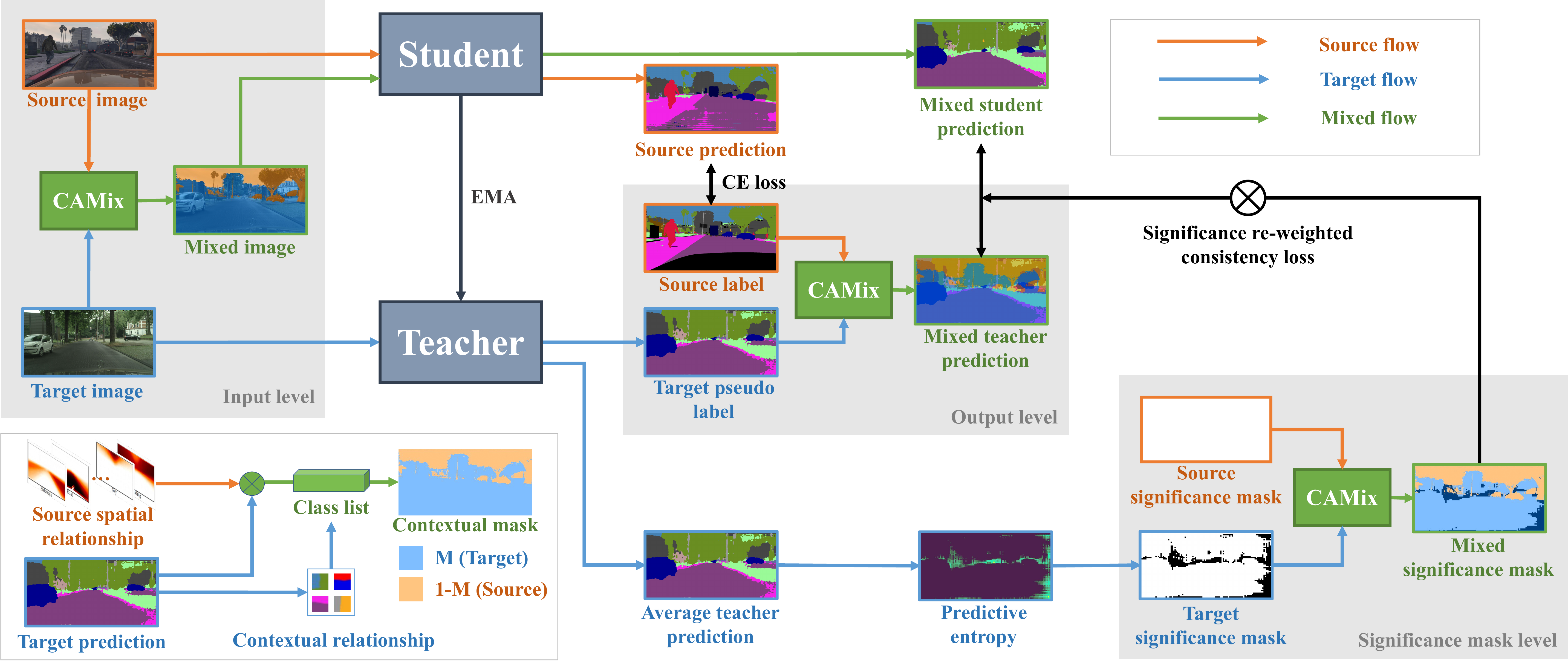}
\caption{Overview of the proposed context-aware mixup (CAMix) architecture. Firstly, we generate a  contextual mask (CMG) by leveraging the spatial distribution of the source domain and the contextual relationship of the target domain. Guided by this mask $M$, we perform context-aware mixup (CAMix) in three levels, \emph{i.e.,} input level, output level and significance mask level. Provided the context knowledge, we design a significance re-weighted consistency (SRC) loss to ease the over-alignment between the mixed student and teacher prediction and alleviate the negative transfer during the adaptation.
}
\label{fig:framework}
\end{figure*}

\section{Related Work}
\noindent \textbf{Unsupervised domain adaptive semantic segmentation.}
Unsupervised domain adaptation (UDA) aims to bridge the domain shifts between the labeled source domain and the unlabeled target domain. This problem has been well-studied in image recognition~\cite{ganin2015dann,deng2020rethinking,zhang2018unsupervised,meng2022slimmable,sun2021joint,zuo2021margin,wang2021confidence,tian2021partial,zhou2022generative}. However, these methods only work on simple and small classification datasets, and may have very limited performance in more challenging and higher-structured tasks, \emph{e.g.,} semantic segmentation. Thus, researching unsupervised domain adaptive semantic segmentation is necessary and significant. 
Many recent methods can be mainly divided into three categories: namely, the input-level adaptation~\cite{Cycada,FDA,BDL,LTIR,CrDoCo,guo2021label,PCEDA,LDR,PIT},  feature-level adaptation~\cite{SIBAN,CBST,CRST,FADA,DADA,lv2020weakly,zhao2022source}, and output-level adaptation~\cite{AdaptSegNet,CLAN,SIM,CLANv2,IntraDA,APODA}. 
However, almost all of them largely overlook the shared context-dependency across domains, leading to less-desired performance. Instead, our method explicitly exploits context dependency across domains as prior domain knowledge for enhancing the adaptability toward the target domain.
Besides, most recent methods~\cite{BDL,FDA,SIM,LTIR} involve many sophisticated sub-components, \emph{e.g.,} computationally-expensive adversarial learning~\cite{AdaptSegNet,CLAN,SIM,CLANv2,IntraDA}, offline self-training~\cite{CBST,CRST,BDL,TPLD,IAST,LSE} and image translation models~\cite{BDL,Cycada,LTIR,LDR,guo2021label}, which are complex and hard to converge, and cannot be trained in an end-to-end manner. In contrast, our method is fully end-to-end trainable and can be easily plugged into existing UDA frameworks,  \emph{e.g.,} DACS~\cite{tranheden2020dacs} and DAFormer~\cite{hoyer2022daformer}.

\noindent \textbf{Domain mixup}: Mixup has been well-studied  in  other communities to improve the robustness of models., \emph{e.g.,} semi-supervised learning~\cite{french2019semi,french2020milking}, and point cloud classification~\cite{zhang2021pointcutmix,chen2020pointmixup}. A few works~\cite{xu2020adversarial,wu2020dual,mao2019virtual} studied cross-domain mixup in UDA. Nevertheless, these methods work well on simple and small classification datasets (\emph{e.g.}  MNIST~\cite{MNIST} and
SVHN~\cite{SVHN}), but can hardly be applied to more challenging tasks, \emph{e.g.,} domain adaptive semantic segmentation. DACS~\cite{tranheden2020dacs} was designed for segmentation and proposed to  mix the source samples with the target ones via ClassMix~\cite{olsson2021classmix}. Besides, BAPA-Net~\cite{liu2021bapa} considered the object boundaries to weigh each pixel for promoting CutMix~\cite{french2019semi}, while little attention has been paid to exploiting explicit contextual dependency as prior knowledge to mitigate the domain gaps.

\noindent \textbf{Consistency regularization}: The key idea of consistency regularization is that the target prediction of the student model and that of the teacher model should be invariant under different perturbations. The teacher model is an exponential moving average (EMA) of the student model, and then the teacher model could transfer the learned knowledge to the student.  Consistency regularization typically appears in Semi-supervised Learning (SSL)~\cite{Mean_teacher} and is recently applied to UDA recently~\cite{french2018self,choi2019self,medical_self_emsembling,olsson2021classmix,tranheden2020dacs,SEANet,zhou2022domain}. For simplicity, we choose \cite{Mean_teacher} as a base framework to realize end-to-end learning. Recent methods~\cite{tranheden2020dacs,olsson2021classmix,liu2021bapa,zhou2022domain} reveal that Cross-Entropy (CE) loss is more suitable than Mean Square Error (MSE) loss and Kullback-Leibler (KL) loss for the segmentation task. Thus, we design our SRC loss as a variant of CE loss.

\noindent \textbf{Uncertainty estimation}: The idea of exploiting prediction uncertainty has been utilized in domain adaptation for classification, \emph{e.g.,} Bayesian classifier~\cite{wen2019bayesian}  and Bayesian discriminator~\cite{kurmi2019attending}. These methods always require an extra discriminator in adversarial training, and can work well on simple and small classification datasets. \textit{Our method differs from these methods in several aspects.} 
At first, we tackle the more challenging task of semantic segmentation rather than image classification, where the uncertainty of dense pixel-wise predictions instead of image-wise prediction needs to be decreased. Secondly, we avoid using adversarial adaptation in uncertainty estimation which tends to be unstable and inaccurate. 
Thirdly, in comparison with the aforementioned approaches, we design significance mask level domain mixup between the target significance mask and the source mask, which enables a more informative entropy-guided mask during the domain mixup.

\begin{figure}[t]
\centering
\includegraphics[scale=1.0]{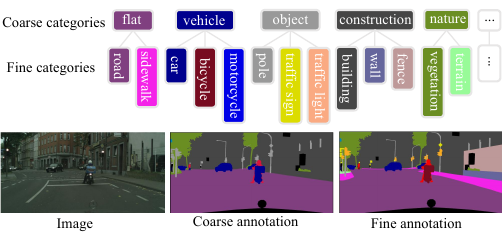}
\caption{
Hierarchical prior contexts given in the Cityscapes~\cite{cordts2016cityscapes} dataset and are shared in all cross-domain scenarios. The top row shows the hierarchy of semantic categories
where multiple fine categories may belong to one coarse category. The bottom row is a street-scene image (left), together with its coarse annotation (middle) and fine annotation (right).
}
\label{fig:context_prior}
\end{figure}

\section{Methodology}

\subsection{Overview and Notations}

Following the UDA protocols~\cite{AdaptSegNet,ADVENT,CBST}, we have access to the source images $X_S \in S$ with their corresponding labels $Y_S$. For the target domain $T$, only unlabeled images $X_T \in T$ are available. Unlike most of the existing UDA methods that overlook the shared context knowledge across domains, we propose a novel context-aware domain mixup (CAMix) to exploit and transfer such cross-domain contexts.

Fig. \ref{fig:framework} shows the overview of our proposed architecture. Firstly, we present a contextual mask generation (CMG) strategy for
mining the prior spatial distribution of the source domain and the contextual relationships of the target domain, thus generating a contextual mask $M$. Guided by this mask $M$, we perform an efficient CAMix on three levels, \emph{i.e.,} input level, output level, and significance mask level (a mask that we define to indicate where the pixels are credible). In particular, the weights of the teacher model $F_{\theta^{\prime}}$ are an exponential moving average (EMA) of the ones of the student model $F_{\theta}$. 
In other words, the proposed CAMix uses the labeled source samples $(X_S, Y_S)$ and unlabeled target samples $X_T$ to synthesize the mixed images $X_M$, the mixed pseudo labels $Y_M$ (Section \ref{sec:inputoutputsampling}), and the corresponding mixed significance masks $U_M$ (Section \ref{sec:siglevelsampling}). We introduce a significance-reweighted consistency loss (SRC) on the significance mask (SigMask) level to alleviate the negative transfer and over-alignment during the online adaptation procedure.

As for other notations,  $\Phi_{t}$ is the model's weight at the $t$-th iteration, $Q$ denotes the spatial prior tensor defined in Section \ref{sec:cmg} and $(h,w)$ means the pixel at $h$ in height and $w$ in width. $H$ and $W$ are the height and width of the image. $L$ denotes the number of stochastic forward passes, $\zeta$ is the predictive entropy, and $R$ is the dynamic threshold defined in Section \ref{sec:siglevelsampling}.
$\boldsymbol{P}_{l}$ is the predicted class scores of the perturbed sample $X_T^l$ at the $l$-th stochastic forward pass, and
$\hat{\boldsymbol{P}}$ is the mean of the predictive probability $\boldsymbol{P}_{l}$ of different forward passes. $P_S$ indicates the predicted scores of the source sample.

\begin{table}[t]

\caption{
Prior knowledge of hierarchical contexts given in the Cityscapes~\cite{cordts2016cityscapes}. We use it to define our meta class group.}
\label{table:meta_class}
\begin{center}
\resizebox{0.38\textwidth}{!}{%

\begin{tabular}{c|c|c}

\toprule
Group & Coarse & Fine \\
\midrule
\multirow{3}{*}{I} & \multirow{3}{*}{ object } & pole\\
  & & traffic sign\\
  & & traffic light \\
\midrule
\multirow{3}{*}{II} & \multirow{3}{*}{human-vehicle} & rider\\
  & & motorcycle\\
  & & bicycle \\
\midrule
\multirow{2}{*}{III} & \multirow{2}{*}{flat} & road\\
  & & sidewalk \\
\midrule
\multirow{3}{*}{IV} & \multirow{3}{*}{construction} & building\\
  & & wall\\
  & & fence \\

\midrule
 \multirow{2}{*}{V } & \multirow{2}{*}{ nature } & vegetation \\
 & & terrain \\

\bottomrule
\end{tabular}
}
\end{center}
\end{table}

\subsection{Contextual Mask Generation}
\label{sec:cmg}
Intuitively, the source and the target domain share similar context dependency between domains. With this in mind, we identify two kinds of semantic contexts as explicit prior domain knowledge for guiding the domain adaptation procedure. The former is prior spatial contexts of the source domain, shown in Fig.~\ref{fig:framework}, and the latter is contextual relationships of the categories in the target domain, shown in Fig.~\ref{fig:context_prior}. 

Regarding that the scenes often have their intrinsic spatial structures, \emph{e.g.,} the sky tends to appear on the top of the image while roads are more likely to appear on the bottom, it is intuitive to explore the spatial relationships of the source domain. Thus, we generate a spatial prior tensor $Q$ with the shape of $C*H*W$ by counting the class frequencies in the source domain.
Each spatial location of $Q$ is a class distribution, and we treat it as prior knowledge to regularize the target prediction: $\hat{F}_{\theta^{\prime}} \leftarrow Q \odot F_{\theta^{\prime}}(X_T)$, where $f_{\theta^{\prime}}(T)$ is the target prediction of the teacher model.

As shown in Fig.~\ref{fig:context_prior}, the object categories of an outdoor scene can be organized by a semantic hierarchy. Cityscapes~\cite{cordts2016cityscapes} dataset gives both the coarse annotation and fine annotation, where multiple fine categories may belong to one coarse category. Strictly following the common UDA protocols that the target training set is unlabeled and only the validation set of the target domain has labels, we use the hierarchical contextual relationship, \emph{i.e.,} the name file of the categories, in the target domain to define the meta class groups of CAMix, as shown in Table \ref{table:meta_class}, for enhancing the adaptability towards the target domain.
\begin{algorithm}[t!]
	\caption{Contextual Mask Generation Algorithm}\label{algorithm 0}
     \KwIn{teacher model $F_{\theta^{\prime}}$, target image $X_T$, spatial matrix $Q$, a meta-class list $m$.}
      \KwOut{contextual mask $M$ for CAMix.}
      $\hat{F}_{\theta^{\prime}} \leftarrow Q \odot F_{\theta^{\prime}}(X_T)$\;
      $\tilde{Y}_{T} \leftarrow \arg \max _{c^{\prime}}\; \hat{f}_{\theta^{\prime}} \left(h, w, c^{\prime}\right)$\;
      $C \leftarrow$ Set of the classes present in $\tilde{Y}_{T}$\;
      $c \leftarrow$ Randomly select $|C| / 2$ classes in $C$\;
      \For{each $k \in c$ }{
         \If{$k \in c$ and $k \in m$}{
            $\tilde{k} \leftarrow $ the semantic-related classes of $k$\;
            \If{$\tilde{k} \in C$}{
            $c.append(\tilde{k})$\;
           }
        }
      } 

      \For{each $h,w$ }{
        $M(h, w)=\left\{\begin{array}{l}1, \text { if } \tilde{Y}_{T}(h, w) \in c \\ 0, \text { otherwise }\end{array}\right.$\
      }
      return $M$\;
\end{algorithm}
To better exploit such contextual relationship for adaptation, \emph{e.g.,} the traffic sign should be beside the pole, our core idea is to find the semantic-related categories of the current class presented in the image  during the domain mixup. In other words, these classes that have contextual relationships to each other can be treated as a meta-class, and then we copy them together from the target images and paste them onto the source images. Our strategy prevents certain semantic categories hanging on an inappropriate context.

Specifically, we first get the spatially-modulated pseudo label: $\tilde{Y}_{T} \leftarrow \arg \max _{c^{\prime}}\hat{F}_{\theta^{\prime}} \left(h, w, c^{\prime}\right)$. Next, we randomly select half of the classes present in the  prediction $\tilde{Y}_{T}$, namely $c$.  After that, we judge whether each category $k \in c$ presented in $\tilde{Y}_{T}$ is in the meta-class list $m$ or not. As shown in Table~\ref{table:meta_class}, the meta-class list $m$ involves several groups of meta-classes, \emph{e.g.,} pole, traffic sign, traffic light are in one group, namely ``object", and bicycle, motorcycle, and rider, are in another group, namely ``human-vehicle", \emph{etc}.
This list is chosen from the prior knowledge of the hierarchical contexts given in the target domain, \emph{i.e.,} Cityscapes~\cite{cordts2016cityscapes}, and  is shared in all experiments. We empirically set $m$ as the combination of group I and group II in all experiments since relatively little context
knowledge is not enough to provide sufficient supervisions for the adaptation, while too much prior knowledge will easily push the learning falling into local optima. More experimental analysis on meta class lists $m$ could be referred to Section~\ref{sec:ablation}.
If $k \in c$, we append the semantic-related classes $\tilde{k}$ of current class $k$ to the current list $c$. 

A binary contextual mask $M$ is then generated by setting the pixels from the final class list $c$ to value $1$ in $M$, and all others to value $0$, which can be formulated as follows: 
\begin{equation}M(h, w)=\left\{\begin{array}{l}
1, \text { if } \tilde{Y}_{T}(h, w) \in c \\
0, \text { otherwise }
\end{array}\right.\end{equation}
where $h \in H, w \in W$, and $H$ and $W$ are the height and width of the image. We iterate each spatial location $(h,w)$ to generate the mask $M$.
\textit{This mask $M$ is then utilized as prior knowledge} to mix the images in the input level, the labels in output level (Section \ref{sec:inputoutputsampling}), and the significance mask  on the significance mask level (Section \ref{sec:siglevelsampling}) between the source domain and the target domain. The whole algorithm of contextual mask generation is described in Algorithm ~\ref{algorithm 0}.

\subsection{Input-level and Output-level Domain Mixup}
\label{sec:inputoutputsampling}
In the \textit{input level}, the image $X_S$ and $X_T$ sampled from the source domain  and target domain are synthesized into $X_{M}$:
\begin{align}
  \label{eq:1}
  \begin{array}{l}
X_{M} =M \odot X_T+(1-M) \odot X_S,
\end{array} 
\end{align}
where $\odot$ denotes element-wise multiplication. 
The weights $\Phi^{'}_{t}$ of the teacher model at training step $t$ are updated by the student's weights $\Phi_{t}$  with a smoothing coefficient $\alpha \in [0,1]$, which can be formulated as follows:
\begin{align}
\label{eq:2}
\Phi^{'}_{t} = \alpha \cdot \Phi^{'}_{t-1} + (1-\alpha) \cdot \Phi_{t} , 
\end{align}
where $\alpha$ is the EMA decay that controls the updating rate.

Regarding the \textit{output level}, the source label $Y_S$ and the target pseudo label $\hat{Y}_{T}=F_{\theta^{\prime}}(X_T)$ are mixed into $Y_M$:
\begin{align}
\label{eq:3}
 Y_{M} =M \odot \hat{Y}_{T} + (1-M) \odot Y_{S}. 
\end{align}

Different from \cite{tranheden2020dacs,olsson2021classmix,hoyer2022daformer}, we mix the images and the corresponding labels in a target-to-source direction rather than the source-to-target direction. In other words, we copy some categories from the target domain and paste them onto the source domain, where we can add our consideration of both spatial relationships and contextual relationships in such a direction. Considering that the target predictions are uncertain without sufficient supervision, these two kinds of context dependencies are more suitable to refine the domain mixup.

\subsection{Significance-mask Level Domain Mixup}
\label{sec:siglevelsampling}
In the significance-mask (SigMask) level domain mixup, we aim to decrease the high uncertainties of the pixel-wise mixed teacher prediction with the guidance of contextual mask $M$ as additional supervisory signals. As a result, we can alleviate the adverse impact, \emph{e.g.,} training instability and early performance degradation, and transfer more reasonable knowledge from the teacher model to the student model.

\noindent \textbf{Stochastic forward passes.}  In particular, following prior works~\cite{choi2019self, zhou2020uncertainty}, we repeat each target image $X_T$ for $L$ copies and inject a random Gaussian noise for each copy. 
Then, for each stochastic forward pass $l$ of the perturbed target sample $X_T^{l}$, we get a set of predicted class scores $\{\boldsymbol{P}_{l}^{(h, w, c)}\}$ at the pixel $(h,w)$ of the $c$-th class. 
Next, we calculate the mean of the predictive probability in $L$ forward passes:
\begin{align}
\label{eq:4}
  \hat{\boldsymbol{P}}^{(h,w,c)} = \frac{1}{L}\sum\limits_{l=1}^{L}\boldsymbol{P}_{l}^{(h, w, c)}(X_T^{l}).
\end{align}
Note that we do not use any dropout layers during stochastic forward passes. The predictive entropy $\zeta$ is calculated as:
\begin{align}
\label{eq:5}
  \zeta^{(h,w)} =-\sum\limits_{c=1}^{C}\hat{\boldsymbol{P}}^{(h,w,c)}\cdot log(\hat{\boldsymbol{P}}^{(h,w,c)}),
\end{align}
where all volumes of pixel-wise entropy $\zeta^{(h,w)}$ form a set: $K=\{\zeta\}_{j=1}^{N}$, and $N$ is the number of pixels in one sample. 

\noindent \textbf{Dynamic threshold.} 
Inspired by the ramp-up function of consistency weight \cite{choi2019self}, a dynamic threshold $R$ is then determined by the predictive entropy rather than the softmax probabilities, which is for filtering out the unreliable pixel-wise mixed teacher predictions. It increases with a lower speed in the early training and a higher speed in the later training:
\begin{align}
\label{eq:6}
    R=\beta+(1-\beta) \cdot e^{\gamma (1-t/t_{max})^2} \cdot K_{sup},
\end{align}
where $t$ denotes the current training step and $t_{max}$ is the  maximum training step. $K_{sup}$ means the upper-bound of the volumes' self-information, which is denoted as: $K_{sup}=sup\{\zeta\}_{j=1}^{N}$.
$\beta$ is the initial state of the dynamic threshold $R$, and $\gamma$ controls the exponential speed of the dynamic threshold.

\noindent \textbf{Significance mask.}
To filter out the unreliable pixel-wise prediction of the mixed teacher predictions, we denote the SigMask $U_T = I(\zeta < R)$ with the help of target predictive entropy $\zeta$ and its dynamic threshold $R$, where $I$ is an indicator function. Only those high-confident pixels where the predictive entropy is lower than the dynamic threshold will remain.

Given the contextual mask $M$ as additional supervisory signals to promote the domain mixup, we perform \textbf{SigMask level} domain mixup. The significance mask of the source domain $U_S$ and the target domain $U_{T}$ are mixed into $U_M$:
\begin{align}
\label{eq:7}
 U_{M}=M \odot U_{T} + (1-M) \odot U_{S},
\end{align}
where $U_{S}$ is a tensor full of 1, because the source labels are provided without uncertainties. And these certain areas do not need to reweigh the consistency loss. Only the uncertain areas in the target $U_{T}$ which is below the dynamic threshold $R$, are set to 0 to reweigh the consistency loss.

\noindent \textbf{Significance-reweighted consistency (SRC) loss.} To encourage the teacher model to transfer more credible knowledge to the student model, we define an SRC loss to penalize the inconsistency between the mixed teacher prediction and the mixed student prediction with the guidance of $U_M$:
\begin{equation}
\label{eq:8}
\mathcal{L}_{con}\left(f_{\theta^{\prime}}, f_{\theta} \right)=
\frac{\sum_{j} \left( U_M \cdot  CE(F_{\theta}(X_M), Y_M) \right)}{\sum_{j} U_M},
\end{equation}
where $F_{\theta^{\prime}}$ and $F_{\theta}$ are the teacher model and the student model, respectively. $CE$ is the abbreviation of the cross-entropy loss. As recent methods~\cite{tranheden2020dacs,olsson2021classmix,liu2021bapa} reveal that Cross-Entropy loss is more suitable than $MSE$ and $KL$ loss for the semantic segmentation task,  we thus design this SRC loss on top of $CE$ loss. 
The pixel-wise SigMask $U_{M}$ is used to reweigh the consistency loss in a weighted averaging manner. In particular, we normalize the loss $\mathcal{L}_{con}$ by the summation of all pixels in the SigMask $U_M$. As a result, we could further alleviate the adverse impacts and the negative transfer, \emph{e.g.,} early performance degradation, during the online adaptation of consistency regularization.

\begin{algorithm}[t!]
	\caption{Context Aware Mixup Algorithm}
	\label{algorithm 1}
	\KwIn{student model $F_{\theta}$, teacher model $F_{\theta^{\prime}}$, source domain $D_S$, target domain $D_T$, total iterations $N$.
	}
	\KwOut{teacher model $F_{\theta^{\prime}}$.}
      Initialize network parameters $\theta$ randomly. \;
      \For{i=1 \KwTo{N}}{
        $X_{S}, Y_{S} \sim \mathcal{D}_{S}$\;
        $X_{T} \sim \mathcal{D}_{T}$\;
        $\hat{Y}_{T} \leftarrow f_{\theta^{\prime}}(X_T)$\;
        $X_M \leftarrow $ Input-level mixup by Eq.\eqref{eq:1}\;
        $\hat{Y}_{S} \leftarrow F_{\theta}\left(X_{S}\right), \hat{Y}_{M} \leftarrow F_{\theta}\left(X_{M}\right)$\;
        $Y_M \leftarrow $ Output-level mixup by Eq.\eqref{eq:3}\;
        $U_T \leftarrow $ Target SigMask by Eq.(\ref{eq:4})$\sim$Eq.\eqref{eq:6} \;
        $U_M \leftarrow $ SigMask-level mixup by Eq.\eqref{eq:7}\;
        $\mathcal{L}_{total} \leftarrow $ Total loss by Eq.(\ref{eq:10})\;
        Compute $\nabla_{\theta} \mathcal{L}_{total}$ by back-propagation\;
        Perform stochastic gradient descent on $\theta$\;
	}
\end{algorithm}

\subsection{End-to-End Training and Inference}

\noindent \textbf{Segmentation loss.} The segmentation loss $L_{seg}$ is a cross-entropy loss for optimizing the source images:
\begin{align}
\label{eq:9}
  \mathcal{L}_{seg} = -\sum\limits_{h=1}^{H}\sum\limits_{w=1}^{W} \sum\limits_{c=1}^{C} Y_{S}^{(h,w,c)}log(P_S^{(h,w,c)}),
\end{align}
where $Y_{S}$ is the ground truth for source images and $P_S = f_{\theta}(X_S)^{(h, w, c)})$ is the segmentation output of source images.

\noindent \textbf{Total loss.} During training, all models on three different levels are jointly trained in an end-to-end manner. The whole framework is optimized by integrating all the loss functions:
\begin{align}
\label{eq:10}
  \mathcal{L}_{total} = \mathcal{L}_{seg}+\lambda_{con} \mathcal{L}_{con},
\end{align}
where $\lambda_{con}$ is the weight of consistency loss. 
, and we use the same adaptive schedule for the weight $\lambda_{con}$ as ~\cite{tranheden2020dacs} in all experiments.
Algorithm \ref{algorithm 1} illustrates the CAMix algorithm of the whole end-to-end training process. 

\noindent
\noindent \textbf{Inference phase:} Since the teacher model is the exponential moving average (EMA) of the student model in the Mean Teacher~\cite{Mean_teacher} architecture, as shown in Eq. \ref{eq:2}, the teacher model always performs slightly better than the student model. Thus, following \cite{Mean_teacher,choi2019self,zhou2020uncertainty,zhou2022domain},  we only use the teacher model to make predictions in the inference phase. 

\subsection{Discussions on differences from 
related work UACR~\cite{zhou2020uncertainty}}
In this subsection, we discuss the differences from the related work UACR~\cite{zhou2020uncertainty} from the following three aspects:

\noindent 
\noindent \textbf{Different motivations:} UACR~\cite{zhou2020uncertainty} focuses on addressing the unreliable guidance of the teacher model in Mean Teacher~\cite{Mean_teacher} architecture and utilizes uncertainty to re-calibrate the teacher predictions. However, UCAR~\cite{zhou2020uncertainty} does not consider exploiting contexts
as explicit prior knowledge for enhancing the adaptability towards the target domain. In contrast, in this work, we observe that almost all existing UDA frameworks largely neglect such context-dependency, which is generally shared across different domains, leading to less-desired performance. From a new perspective, our goal is to exploit this important clue of context-dependency as explicit prior knowledge to promote the domain mixup, which is different from \cite{zhou2020uncertainty}.

\noindent 
\noindent \textbf{Different frameworks:} Although our CAMix also performs stochastic forward passes to estimate uncertainty, our main contribution is not it but a CAMix framework that performs context-aware mixup in three different levels, which selectively leverages the spatial distribution of the source domain and the contextual relationship of the target domain.
Besides, UCAR~\cite{zhou2020uncertainty} needs to utilize the image translation model, \emph{e.g.,} CycleGAN~\cite{CycleGAN2017}, to stylize the source domain to the intermediate domain with target styles, which requires two-stage training for adaptation. In contrast, our CAMix framework can be trained in a fully end-to-end manner, which largely simplifies the training procedure and is more practical in real-world applications. In the experimental part, we demonstrate that our CAMix outperforms UCAR~\cite{zhou2020uncertainty} by a large margin in two benchmarks, shown in Table~\ref{table:gtav} and Table~\ref{table:synthia}.

\noindent 
\noindent \textbf{Different constraints:} As for the consistency loss, UCAR~\cite{zhou2020uncertainty} utilized both the uncertainty mask and classdrop mask of the target images for reweighing the original teacher prediction without any mixing operations. In particular, UCAR~\cite{zhou2020uncertainty} utilized a ClassOut strategy to ensure the model will produce consistent predictions under the ClassDrop perturbations. In contrast, in this work, we mix the source significance mask and the target mask via the proposed CAMix, and then utilize the mixed significance mask to reweigh the mixed teacher predictions, which is different from \cite{zhou2020uncertainty}.

\begin{table*}{}
\caption{
Comparison results (mIoU) with state-of-the-art methods from GTAV to Cityscapes. }
\label{table:gtav}
\centering
\resizebox{\textwidth}{!}{%
\begin{tabular}{c|c|ccccccccccccccccccc|c}
\toprule
Method& Venue                 & \begin{turn}{90}road\end{turn} & \begin{turn}{90}sidewalk\end{turn} & \begin{turn}{90}building\end{turn} & \begin{turn}{90}wall\end{turn} & \begin{turn}{90}fence\end{turn} & \begin{turn}{90}pole\end{turn} & \begin{turn}{90}light\end{turn} & \begin{turn}{90}sign\end{turn} & \begin{turn}{90}vegetation\end{turn} & \begin{turn}{90}terrain\end{turn} & \begin{turn}{90}sky\end{turn} & \begin{turn}{90}person\end{turn} & \begin{turn}{90}rider\end{turn} & \begin{turn}{90}car\end{turn} & \begin{turn}{90}truck\end{turn} & \begin{turn}{90}bus\end{turn} & \begin{turn}{90}train\end{turn} & \begin{turn}{90}motocycle\end{turn} & \begin{turn}{90}bike\end{turn} & \begin{turn}{90}\textbf{mIoU}\end{turn}  \\ 
\toprule
SIBAN~\cite{SIBAN}&ICCV'19 &88.5 &35.4 &79.5 &26.3 &24.3 &28.5 &32.5 &18.3 &81.2 &40.0 &76.5 &58.1 &25.8 &82.6 &30.3 &34.4 &3.4 &21.6 &21.5 &42.6\\
BDL~\cite{BDL}&CVPR'19 &91.0 &44.7 &84.2 &34.6 &27.6 &30.2 &36.0 &36.0 &85.0 &43.6 &83.0 &58.6 &31.6 &83.3 &35.3 &49.7 &3.3 &28.8 &35.6 &48.5 \\
APODA~\cite{APODA}&AAAI'20  &85.6 &32.8 &79.0 &29.5 &25.5 &26.8 &34.6 &19.9 &83.7 &40.6 &77.9 &59.2 &28.3 &84.6 &34.6 &49.2 &8.0 &32.6 &39.6 &45.9\\
IntraDA~\cite{IntraDA}&CVPR'20 &90.6 &37.1 &82.6 &30.1 &19.1 &29.5 &32.4 &20.6 &85.7 &40.5 &79.7 &58.7 &31.1 &86.3 &31.5 &48.3 &0.0 &30.2 &35.8 &46.3 \\
SIM~\cite{SIM}&CVPR'20 &90.6 &44.7 &84.8 &34.3 &28.7 &31.6 &35.0 &37.6 &84.7 &43.3 &85.3 &57.0 &31.5 &83.8 &42.6 &48.5 &1.9 &30.4 &39.0 & 49.2 \\
LTIR~\cite{LTIR}&CVPR'20 &92.9 &55.0 &85.3 &34.2 &31.1 &34.9 &40.7 &34.0 &85.2 &40.1 &87.1 &61.0 &31.1 &82.5 &32.3 &42.9 &0.3 &36.4 &46.1 & 50.2 \\
FDA~\cite{FDA}&CVPR'20 &92.5 &53.3 &82.4 &26.5 &27.6 &36.4 &40.6 &38.9 &82.3 &39.8 &78.0 &62.6 &34.4 &84.9 &34.1 &53.1 &16.9 &27.7 &46.4 &50.5 \\
PCEDA~\cite{PCEDA}&CVPR'20 &91.0 &49.2 &85.6 &37.2 &29.7 &33.7 &38.1 &39.2 &85.4 &35.4 &85.1 &61.1 &32.8 &84.1 &45.6 &46.9 &0.0 &34.2 &44.5 &50.5\\

LSE~\cite{LSE}&ECCV'20 &90.2 &40.0 &83.5 &31.9 &26.4 &32.6 &38.7 &37.5 &81.0 &34.2 &84.6 &61.6 &33.4 &82.5 &32.8 &45.9 &6.7 &29.1 &30.6 &47.5\\
WLabel~\cite{WLabel}&ECCV'20 &91.6 &47.4 &84.0 &30.4 &28.3 &31.4 &37.4 &35.4 &83.9 &38.3 &83.9 &61.2 &28.2 &83.7 &28.8 &41.3 &8.8 & 24.7 &46.4 &48.2 \\
CrCDA~\cite{CrCDA}&ECCV'20 &92.4 &55.3 &82.3 &31.2 &29.1 &32.5 &33.2 &35.6 &83.5 &34.8 &84.2 &58.9 &32.2 &84.7 &40.6 &46.1 &2.1 &31.1 &32.7 &48.6 \\
FADA~\cite{FADA}&ECCV'20 &92.5 &47.5 &85.1 &37.6 &32.8 &33.4 &33.8 &18.4 &85.3 &37.7 &83.5 &63.2 &39.7 &87.5 &32.9 &47.8 &1.6 &34.9 &39.5 & 49.2 \\
LDR~\cite{LDR}&ECCV'20 &90.8 &41.4 &84.7 &35.1 &27.5 &31.2 &38.0 &32.8 &85.6 &42.1 &84.9 &59.6 &34.4 &85.0 &42.8 &52.7 &3.4 &30.9 &38.1 &49.5\\
CCM~\cite{CCM}&ECCV'20 &93.5 &57.6 &84.6 &39.3 &24.1 &25.2 &35.0 &17.3 &85.0 &40.6 &86.5 &58.7 &28.7 &85.8 &49.0 &56.4 &5.4 &31.9 &43.2 &49.9 \\
CD-SAM~\cite{yang2021context} &WACV'21 &91.3 &46.0 &84.5 &34.4 &29.7 &32.6 &35.8 &36.4 &84.5 &43.2 &83.0 &60.0 &32.2 &83.2 &35.0 &46.7 &0.0 &33.7 &42.2 &49.2 \\
ASA~\cite{ASA} & TIP'21 &89.2 &27.8 &81.3 &25.3 &22.7 &28.7 &36.5 &19.6 &83.8 &31.4 &77.1 &59.2 &29.8 &84.3 &33.2 &45.6 &16.9 &34.5 &30.8 &45.1 \\
CLAN~\cite{CLANv2} & TPAMI'21 &88.7 &35.5 &80.3 &27.5 &25.0 &29.3 &36.4 &28.1 &84.5 &37.0 &76.6 &58.4 &29.7 &81.2 &38.8 &40.9 &5.6 &32.9 &28.8 &45.5\\
DAST~\cite{DAST}&AAAI'21 &92.2 &49.0 &84.3 &36.5 &28.9 &33.9 &38.8 &28.4 &84.9 &41.6 &83.2 &60.0 &28.7 &87.2 &45.0 &45.3 &7.4 &33.8 &32.8 &49.6 \\
BiMaL~\cite{truong2021bimal} &ICCV'21 & 91.2 & 39.6 & 82.7 & 29.4 & 25.2 & 29.6 & 34.3 & 25.5 & 85.4 & 44.0 & 80.8 & 59.7 & 30.4 & 86.6 & 38.5 & 47.6 & 1.2 & 34.0 & 36.8 & 47.3\\
UncerDA~\cite{wang2021uncertainty} &ICCV'21 &90.5 &38.7 &86.5 &41.1 &32.9 &40.5 &48.2 &42.1 &86.5 &36.8 &84.2 &64.5 &38.1 &87.2 &34.8 &50.4 &0.2 &41.8 &54.6 &52.6 \\
DPL-Dual~\cite{cheng2021dual} &ICCV'21 &92.8 &54.4 &86.2 &41.6 &32.7 &36.4 &49.0 &34.0 &85.8 &41.3 &86.0 &63.2 &34.2 &87.2 &39.3 &44.5 &18.7 &42.6 &43.1 &53.3 \\
RPLR~\cite{li2022featurere}& TPAMI'22 &92.3  &52.3  &84.8  &34.7  &29.7  &32.6  &36.7  &32.7  &83.2  &42.5  &81.5  &60.6  &33.3  &85.0  &44.2  &48.0  &3.8  &35.7  &37.3  &50.1 \\ 
UACR~\cite{zhou2020uncertainty} & CVIU'22 &91.3 &48.6 &85.5 &35.8 &31.4 &36.7 &37.5 &36.8 &86.3 &40.3 &85.7 &64.3 &31.1 &87.7 &36.7 &44.9 &15.9 &38.9 &55.4 &51.9\\
\midrule
DACS~\cite{tranheden2020dacs}& WACV'21 &89.9 &39.7 &87.9 &30.7 &39.5 &38.5 &46.4 &52.8 &88.0 &44.0 &88.8 &67.2 &35.8 &84.5 &45.7 &50.2 &0.0 &27.3 &34.0 &52.1\\
Ours (w DACS~\cite{tranheden2020dacs})& - &93.3 &58.2 &86.5 &36.8 &31.5 &36.4 &35.0 &43.5 &87.2 &44.6 &88.1 &65.0 &24.7 &89.7 &46.9 &56.8 &27.5 &41.1 &56.0 &55.2\\
\midrule
DAFormer~\cite{hoyer2022daformer} &CVPR'22 & 95.7 & 70.2 & 89.4 & 53.5 & 48.1 & 49.6 & 55.8 & 59.4 & 89.9 & 47.9 & \textbf{92.5} & \textbf{72.2} & \textbf{44.7} & 92.3 & 74.5 & 78.2 & 65.1 & 55.9 & 61.8 & 68.3\\
Ours (w DAFormer~\cite{hoyer2022daformer}) & - &\textbf{96.0} &\textbf{73.1} &\textbf{89.5} &\textbf{53.9} &\textbf{50.8} &\textbf{51.7} &\textbf{58.7} &\textbf{64.9} &\textbf{90.0} &\textbf{51.2} &92.2 &71.8 &44.0 &\textbf{92.8} &\textbf{78.7} &\textbf{82.3} &\textbf{70.9} &\textbf{54.1} &\textbf{64.3} &\textbf{70.0}\\
\bottomrule
 \end{tabular}}
\end{table*}
\section{Experiments}
In this section, we first describe the experimental setup in Section~\ref{sec:datasets} and implementation details in Section~\ref{sec:implementation details}. Then, we demonstrate the effectiveness of our framework on two widely-used UDA benchmarks, \emph{i.e.,}  GTAV~\cite{stephan2016gtav}
$\rightarrow$ Cityscapes~\cite{cordts2016cityscapes}, and SYNTHIA~\cite{ros2016synthia} $\rightarrow$ Cityscapes~\cite{cordts2016cityscapes}. Finally, we provide extensive ablation studies with analysis (Section~\ref{sec:ablation}) and visualizations (Section~\ref{sec:visualization}) to
reveal the contribution of each component of our proposed method.

\subsection{Datasets}
\label{sec:datasets}
Following common UDA protocols \cite{FCN_wild, AdaptSegNet}, we use the labeled synthetic dataset, \emph{i.e.,} GTAV~\cite{stephan2016gtav} and SYNTHA~\cite{ros2016synthia}, as the source domain, and the unlabeled real dataset \emph{i.e.,} Cityscapes~\cite{cordts2016cityscapes} as the target domain. 

\noindent \textbf{Cityscapes}~\cite{cordts2016cityscapes} is a dataset focused on autonomous driving,
which consists of 2,975 images in the training set and 500 images in the validation set. The images have
a fixed spatial resolution of 2048 $\times$ 1024 pixels. 
Following common practice, we trained the model on the unlabeled training set and report our results on the validation set.\\
\textbf{GTAV}~\cite{stephan2016gtav} is a synthetic dataset including 24,966 photo-realistic images rendered by the gaming engine Grand Theft Auto V (GTAV).
The semantic categories are compatible between the two datasets. We used all the 19 official training classes in our experiments.\\
\textbf{SYNTHIA}~\cite{ros2016synthia} is a synthetic dataset composed of 9,400
annotated images with the resolution of 1280 $\times$ 960.
It also has semantically compatible annotations with
Cityscapes. Following prior works~\cite{CrossCity,CDA,ROAD},
we use the SYNTHIA-RAND-CITYSCAPES subset~\cite{ros2016synthia} as our training set.

\subsection{Implementation Details}
\label{sec:implementation details}
Following common UDA protocols~\cite{BDL,CLAN}, when the source domain is GTAV~\cite{stephan2016gtav}, we resize all images to $1280 \times 720$; when the source domain is SYNTHIA~\cite{ros2016synthia}, we resize all images to $1280 \times 760$. Then, both the source and target images are randomly cropped to $512 \times 512$. To demonstrate the effectiveness, we implement our method in two popular network architectures, \emph{i.e.,} DeepLabV2~\cite{chen2018deeplab} and SegFormer~\cite{xie2021segformer}.
\begin{table*}[t]
\caption{
Comparison results (mIoU) with state-of-the-art methods from SYNTHIA to Cityscapes.}
\label{table:synthia}
\centering
\resizebox{\textwidth}{!}{%
\begin{tabular}{c|c|ccccccccccccc|c} 
\toprule
Method & Venue               &  \begin{turn}{90}road\end{turn} & \begin{turn}{90}sidewalk\end{turn} & \begin{turn}{90}building\end{turn} & \begin{turn}{90}light\end{turn} & \begin{turn}{90}sign\end{turn} & \begin{turn}{90}vegetation\end{turn} & \begin{turn}{90}sky\end{turn} & \begin{turn}{90}person\end{turn} & \begin{turn}{90}rider\end{turn} & \begin{turn}{90}car\end{turn} & \begin{turn}{90}bus\end{turn} & \begin{turn}{90}motocycle\end{turn} & \begin{turn}{90}bike\end{turn} & \begin{turn}{90}\textbf{mIoU$_{13}$}\end{turn}  \\ 
\toprule
SIBAN~\cite{SIBAN}&ICCV'19 &82.5 &24.0 &79.4 &16.5 &12.7 &79.2 &82.8 &58.3 &18.0 &79.3 &25.3 &17.6 &25.9 &46.3 \\
DADA~\cite{DADA}&ICCV'19 &89.2 &44.8 &81.4 &8.6 &11.1 &81.8 &84.0 &54.7 &19.3 &79.7 &40.7 &14.0 &38.8  &49.8 \\
BDL~\cite{BDL}&CVPR'19 &86.0 &46.7 &80.3 &14.1 &11.6 &79.2 &81.3 &54.1 &27.9 &73.7 &42.2 &25.7 &45.3  &51.4 \\
APODA~\cite{APODA}&AAAI'20 &86.4 &41.3 &79.3 &22.6 &17.3 &80.3 &81.6 &56.9 &21.0 &84.1 &49.1 &24.6 &45.7  &53.1 \\
IntraDA~\cite{IntraDA}&CVPR'20 &84.3 &37.7 &79.5 &9.2 &8.4 &80.0 &84.1 &57.2 &23.0 &78.0 &38.1 &20.3 &36.5 &48.9 \\
LTIR~\cite{LTIR}&CVPR'20 &92.6 &53.2 &79.2 &1.6 &7.5 &78.6 &84.4 &52.6 &20.0 &82.1 &34.8 &14.6 &39.4  &49.3 \\
SIM~\cite{SIM}&CVPR'20 &83.0 &44.0 &80.3 & 17.1 &15.8 &80.5 &81.8 &59.9 &33.1 &70.2 &37.3 &28.5 &45.8  &52.1 \\
FDA~\cite{FDA}&CVPR'20 & 79.3 &35.0 &73.2 &19.9 &24.0 &61.7 &82.6 &61.4 &31.1 &83.9 &40.8 &38.4 &51.1  &52.5\\
LSE~\cite{LSE}&ECCV'20 &82.9 &43.1 &78.1 &9.1 &14.4 &77.0 &83.5 &58.1 &25.9 &71.9 &38.0 &29.4 &31.2  &49.4 \\
CrCDA~\cite{CrCDA}&ECCV'20 &86.2 &44.9 &79.5 &9.4 &11.8 &78.6 &86.5 &57.2 &26.1 &76.8 &39.9 &21.5 &32.1  &50.0 \\
WLabel~\cite{WLabel}&ECCV'20 &92.0 &53.5 &80.9 &3.8 &6.0 &81.6 &84.4 &60.8 &24.4 &80.5 &39.0 &26.0 &41.7  &51.9 \\
CCM~\cite{CCM}&ECCV'20 &79.6 &36.4 &80.6 &22.4 &14.9 &81.8 &77.4 &56.8 &25.9 &80.7 &45.3 &29.9 &52.0  &52.9 \\
LDR~\cite{LDR}&ECCV'20 &85.1 &44.5 &81.0 &16.4 &15.2 &80.1 &84.8 &59.4 &31.9 &73.2 &41.0 &32.6 &44.7 &53.1 \\
CD-SAM~\cite{yang2021context} &WACV'21 &82.5 &42.2 &81.3 &18.3 &15.9 &80.6 &83.5 &61.4 &33.2 &72.9 &39.3 &26.6 &43.9 &52.4\\
CLAN~\cite{CLANv2} &TPAMI'21 &82.7 &37.2 &81.5 &17.1 &13.1 &81.2 &83.3 &55.5 &22.1 &76.6 &30.1 &23.5 &30.7 &48.8\\
ASA~\cite{ASA} &TIP'21 &91.2 &48.5 &80.4 &5.5 &5.2 &79.5 &83.6 &56.4 &21.9 &80.3 &36.2 &20.0 &32.9 &49.3\\
DAST~\cite{DAST}&AAAI'21 &87.1 &44.5 &82.3 &13.9 &13.1
&81.6 &86.0 &60.3 &25.1 &83.1 &40.1 &24.4 &40.5 &52.5\\
BiMaL~\cite{truong2021bimal} &ICCV'21 & 92.8 & 51.5 & 81.5  & 17.6 & 15.9 & 82.4 & 84.6 & 55.9 & 22.3 & 85.7 & 44.5 & 24.6 & 38.8 & 53.7 \\
UncerDA~\cite{wang2021uncertainty} &ICCV'21 &79.4 &34.6 &83.5 &32.1 &26.9 &78.8 &79.6 &66.6 &30.3 &86.1 &36.6 &19.5 &56.9 &54.6 \\
DPL-Dual~\cite{cheng2021dual} &ICCV'21 &87.5  &45.7  &82.8 &22.0  &20.1  &83.1  &86.0  &56.6  &21.9  &83.1  &40.3  &29.8  &45.7 &54.2 \\
RPLR~\cite{li2022featurere}& TPAMI'22 &81.5 &36.7  &78.6  &20.7  &23.6  &79.1  &83.4  &57.6  &30.4  &78.5  &38.3  &24.7  &48.4  &52.4\\
UACR~\cite{zhou2020uncertainty} &CVIU'22 &85.5 &42.5 &83.0 &20.9 &25.5 &82.5 &88.0 &63.2 &31.8 &86.5 &41.2 &25.9 &50.7  &55.9\\
\midrule
DACS~\cite{tranheden2020dacs} & WACV'21 &80.6 &25.1 &81.9 &22.7 &24.0 &83.7 &90.8 &67.6 &38.3 &82.9 &38.9 &28.5 &47.6 &54.8 \\
Ours (w DACS~\cite{tranheden2020dacs}) &- &91.8 &54.9 &83.6 &23.0 &29.0 &83.8 &87.1 &65.0 &26.4 &85.5 &55.1 &36.8 &54.1  &59.7\\
\midrule
DAFormer~\cite{hoyer2022daformer} & CVPR'22 &84.5 &40.7 &88.4 &55.0 &54.6 &86.0 &89.8 &\textbf{73.2} &48.2 &\textbf{87.2} &53.2 &53.9 &61.7 &67.4\\
Ours (w DAFormer~\cite{hoyer2022daformer}) &- & 
\textbf{87.4} & \textbf{47.5} & \textbf{88.8} & \textbf{55.2} & \textbf{55.4} & \textbf{87.0} & \textbf{91.7} & 72.0 & \textbf{49.3} & 86.9 & \textbf{57.0} & \textbf{57.5} & \textbf{63.6} & \textbf{69.2} \\
\bottomrule
\end{tabular}}
\end{table*}
\noindent
\textbf{Implementation Details with DeepLabV2~\cite{chen2018deeplab}:}
Following the widely used implementation protocol in previous works~\cite{tranheden2020dacs,liu2021bapa,BDL,AdaptSegNet,FDA}, we employ DeepLabV2~\cite{chen2018deeplab} with ResNet 101 backbone~\cite{he2016deep}. Following \cite{tranheden2020dacs,olsson2021classmix,liu2021bapa}, the backbone is pre-trained on ImageNet~\cite{deng2009imagenet} and MSCOCO~\cite{COCO}. For the DeepLabV2 network, we use Adam~\cite{kingma2014adam} as the optimizer. The initial learning rate is  $2.5 \times 10 ^{-4}$ which is then decreased using polynomial decay with an exponent of $0.9$. The weight decay is  $5 \times 10 ^{-5}$ and the momentum is $0.9$. We use the same data augmentation as DACS~\cite{tranheden2020dacs}, \emph{i.e.,} color jittering and Gaussian blurring.  Our method is implemented in Pytorch on a single NVIDIA Tesla V100, and we train the model for 250K iterations with an early stop setting. 

\noindent
\textbf{Implementation Details with SegFormer~\cite{xie2021segformer}:}
Following DAFormer~\cite{hoyer2022daformer}, our basic network is based on SegFormer~\cite{xie2021segformer}, which consists of an MiT-B5 encoder~\cite{xie2021segformer} and a context-aware feature fusion decoder. All encoders are pretrained on ImageNet-1k dataset~\cite{deng2009imagenet}. 
We use AdamW~\cite{loshchilov2018decoupled} as the optimizer with a learning rate of $\eta_\mathit{base} {=} 6 {\times} 10^{-5}$ for the encoder and $ 6 {\times} 10^{-4}$ for the decoder, a weight decay of $0.01$, linear learning rate warmup with $t_\mathit{warm}{=}1.5$k, and linear decay afterwards. We use the same DACS~\cite{tranheden2020dacs} data augmentation and set $\alpha{=}0.99$. We train the model for 90K iterations on a single NVIDIA Tesla V100.

\noindent
\textbf{More Details of CAMix:}
Following prior works~\cite{kim2019self,zhou2020uncertainty}, we perform $N=8$ times of stochastic forward passes in our SigMask-level CAMix. Besides, we use the same adaptive schedule as previous consistency regularization works, \emph{e.g,} CutMix~\cite{french2019semi} and DACS~\cite{tranheden2020dacs} for the consistency weight $\lambda_{con}$. As suggested by~\cite{kim2019self,zhou2020uncertainty}, we use the same hyper-parameters of  $\beta = 0.75, \gamma = -5$ by default in Eq. \ref{eq:6} in all experiments.

\subsection{Comparison with the State-of-the-Art Methods}
\label{sec:comparison_to_sota}
Table~\ref{table:gtav} and Table~\ref{table:synthia} present the comparison results with the state-of-the-art methods on two challenging UDA tasks: ``GTAV $\rightarrow$ Cityscapes'' and ``SYNTHIA $\rightarrow$ Cityscapes''. As we can see, our proposed method outperforms these competitors by a large margin with two different baselines methods, \emph{e.g.,} DACS~\cite{tranheden2020dacs} and DAFormer~\cite{hoyer2022daformer}. In particular, our method (w DACS~\cite{tranheden2020dacs}) is superior to the DACS~\cite{tranheden2020dacs} baseline by $3.1\%$ and $4.9\%$ of mIoU in these two benchmarks, and our method (w DAFormer~\cite{hoyer2022daformer}) achieves improvements of $1.7\%$ and $1.8\%$ of mIoU compared to DAFormer~\cite{hoyer2022daformer} in these two datasets.

Specifically, most recent UDA approaches perform adversarial learning, \emph{e.g.,} APODA~\cite{APODA}, IntraDA~\cite{IntraDA}, WLabel~\cite{WLabel}, FADA~\cite{FADA} and DADA~\cite{DADA}, and they need to carefully tune the optimization procedure for min-max problems through a domain discriminator. However, such domain discriminators tend to be unstable and inaccurate. Instead, our method does not require maintaining an extra discriminator during the domain adaptation process, and we outperform these approaches by more than $6\%$ with DACS~\cite{tranheden2020dacs} and $20\%$ with DAFormer~\cite{hoyer2022daformer}. 
To alleviate the adversarial feature misalignment and stabilize the training of the discriminator during the adversarial adaptation, SIBAN~\cite{SIBAN} presents a significance-aware module to detect the channel-wise significance for each pixel-level feature, and for weighting the information bottleneck loss. In contrast, we do not rely on additional models, \emph{e.g.,} the significance-aware module in ~\cite{SIBAN}, for significance reweighting, and simply use Eq. \ref{eq:4} $\sim$ Eq. \ref{eq:7} to compute the significance mask. Besides, given the contextual mask $M$ as additional supervisory signals to promote the domain mixup, our SigMask is used to reweigh the consistency loss, which has different focus from~\cite{SIBAN}. Table~\ref{table:gtav} and Table~\ref{table:synthia} demonstrate that our method is superior to SIBAN~\cite{SIBAN} by a large margin.

\begin{table*}[t]
\centering
\caption{
Comparisons with related domain mixup methods using different basic architectures from GTAV to Cityscapes.}
\label{table:comparison_mixup}
\subfloat[Comparison results using DeeplabV2~\cite{chen2018deeplab} as basic architecture.
\label{table:comparison_mixup:deeplabv2}
]{
\resizebox{0.45\textwidth}{!}{%
\begin{tabular}{l|c|c} \toprule
method (w Deeplab-v2~\cite{chen2018deeplab}) & mIoU (\%) & Gain (\%)\\
\midrule
Mean Teacher~\cite{Mean_teacher} & 43.1 & --\\
+ CowMix~\cite{french2020milking} & 48.3 & +5.2\\
+ CutMix~\cite{french2019semi} & 48.7 & +5.6\\
+ DACS~\cite{tranheden2020dacs} & 52.1 & +9.0\\
+ iDACS~\cite{tranheden2020dacs} & 51.5 & +8.4\\
\midrule
+ Ours (CAMix) & \textbf{55.2} & +\textbf{12.1}\\
\bottomrule
\end{tabular}
}
}
\quad
\quad
\quad
\subfloat[Comparison results using SegFormer~\cite{xie2021segformer} as basic architecture.
\label{table:comparison_mixup:segformer}
]{
\resizebox{0.45\textwidth}{!}{%
\begin{tabular}{l|c|c} \toprule
method (w SegFormer~\cite{xie2021segformer}) & mIoU (\%) & Gain (\%)\\
\midrule
Mean Teacher~\cite{Mean_teacher} & 51.6 & --\\
+ CowMix~\cite{french2020milking} & 58.9 & +7.3\\
+ CutMix~\cite{french2019semi} & 58.7 & +7.1\\
+ DAFormer~\cite{hoyer2022daformer} & 68.3 & +16.7\\
+ iDAFormer~\cite{hoyer2022daformer} & 62.4 & +10.8\\
\midrule
+ Ours (CAMix)  & \textbf{70.0} & +\textbf{18.4}\\
\bottomrule
\end{tabular}
}
}
\vspace{-3mm}
\end{table*}
\begin{table*}[t]
\centering
\caption{
Ablation study of each component and each level in CAMix.}

\label{table:ablation_component_level}
\subfloat[Ablation study of each component in CAMix.
\label{table:ablation_component_camix}
]{
\resizebox{0.38\textwidth}{!}{%
\begin{tabular}{cccc|cc} \toprule
Baseline~\cite{tranheden2020dacs} & SP & CR & SRC & mIoU \\
\midrule
$\surd$  &  &  & & 51.5 \\
$\surd$  & $\surd$ & &  & 53.1\\
$\surd$  & $\surd$ & $\surd$ &  & 54.5\\
$\surd$  & $\surd$  & $\surd$ & $\surd$  & 55.2 \\
\bottomrule
\end{tabular}
}
}
\quad
\quad
\quad
\quad
\quad
\subfloat[Ablation study of each level in CAMix.
\label{table:ablation_level_camix}
]{
\resizebox{0.43\textwidth}{!}{%
\begin{tabular}{ccc|cc} \toprule
\thead{Mean \\ Teacher} & In-Out & SigMask& \thead{mIoU \\ (GTAV)} & \thead{mIoU$_{13}$ \\ (SYN)}\\
\midrule
$\surd$  &  &  & 43.1 & 45.9\\
$\surd$  & $\surd$  &  & 54.5 & 59.0\\
$\surd$  & $\surd$  & $\surd$  & 55.2 &59.7\\
\bottomrule
\end{tabular}
}
}
\vspace{-5mm}
\end{table*}

In contrast to the offline self-training methods that need to fine-tune the models in many rounds, \emph{e.g.,} CRST~\cite{CRST}, LSE~\cite{LSE}, CCM~\cite{CCM}, our whole framework can be trained in a fully end-to-end manner. Benefiting from the online consistency regularization with our specially-designed components, our approach significantly outperforms them by at least $5\%$ with DACS~\cite{tranheden2020dacs} and $16\%$ with DAFormer~\cite{hoyer2022daformer}. 

Compared to the methods which require an image-to-image (I2I) translation or style transfer algorithm to filter out the domain-specific texture or style information, \emph{e.g.,} BDL~\cite{BDL}, 
LDR~\cite{LDR}, LTIR~\cite{LTIR}, FDA~\cite{FDA} and PCEDA~\cite{PCEDA}, our context-aware domain mixup does not require any style/spectral transfer algorithms or deep neural networks for I2I translation. Our CAMix (w DACS~\cite{tranheden2020dacs}) is simple and works very well, and it surpasses the translation-based methods by around $5\% \sim 8\%$. 

CrCDA~\cite{CrCDA} learned and enforced the prototypical local contextual relations in the feature space, and similarly, CD-SAM~\cite{yang2021context} exploits contexts implicitly in the feature space, while the visual cues of context knowledge tend to be lost. Moreover, both of the learning~\cite{CrCDA,yang2021context} do not \textit{explicitly} exploit the cross-domain contexts in the image space and cannot be trained end-to-end. In contrast, our CAMix explicitly explores the contexts in the image space rather than the feature space, and our architecture can be trained end-to-end. Our approach (w DACS~\cite{tranheden2020dacs}) outperforms the CrCDA~\cite{CrCDA}  by $6.6\%$ and $9.7\%$ in two benchmarks, respectively.

Compared to resampling-based methods~\cite{wang2021uncertainty,hoyer2022daformer} that aim to remedy the class imbalance issue in UDA, our method (w DACS~\cite{tranheden2020dacs}) outperforms the UncerDA~\cite{wang2021uncertainty} that uses the soft-balance sampling by $2.6\%$ in GTAV~\cite{stephan2016gtav} $\rightarrow$ Cityscapes~\cite{cordts2016cityscapes} and $5.1\%$ in SYNTHIA~\cite{ros2016synthia} $\rightarrow$ Cityscapes~\cite{cordts2016cityscapes}. Besides, our approach (w DAFormer~\cite{hoyer2022daformer}) is superior to DAFormer~\cite{hoyer2022daformer} that utilizes the rare class sampling (RCS) by $1.7\%$ in GTAV~\cite{stephan2016gtav} $\rightarrow$ Cityscapes~\cite{cordts2016cityscapes} and $1.8\%$ in SYNTHIA~\cite{ros2016synthia} $\rightarrow$ Cityscapes~\cite{cordts2016cityscapes}. 
Taking a closer look at per-category performance in Table~\ref{table:gtav} and Table~\ref{table:synthia}, our approach achieves the highest IoU on most categories, \emph{e.g.,} motorcycle, bicycle, traffic sign, \emph{etc}, and is superior to those resampling-based methods in most categories. This phenomenon reveals the effectiveness of CAMix among different classes during the domain adaptation process.

\subsection{Comparison with related Domain Mixup methods}

As shown in Table~\ref{table:comparison_mixup}, we present the comparison results with the existing domain mixup algorithms on GTAV~\cite{stephan2016gtav} $\rightarrow$ Cityscapes~\cite{cordts2016cityscapes}.  All existing domain mixup algorithms are implemented under the same settings and we choose the Mean Teacher~\cite{Mean_teacher} as our baseline in this experiment.
CowMix~\cite{french2020milking}, CutMix~\cite{french2019semi} are proposed for semi-supervised learning (SSL), and we adapt them to the UDA task, which mixes the source domain image and the target domain image. Besides, we implement the existing cross-domain mixup method, \emph{e.g.,} DACS~\cite{tranheden2020dacs} and inverse DACS.
The former DACS means using ClassMix to copy the source categories and paste them onto the target, and the latter Inverse DACS (iDACS)~\cite{tranheden2020dacs} uses a target-to-source direction. Similarly, DAFormer~\cite{hoyer2022daformer} and inverse DAFormer (iDAFormer)~\cite{hoyer2022daformer} use the source-to-target and target-to-source directions, respectively, during the domain mixup. Note that all experiments of our method are based on the iDACS~\cite{tranheden2020dacs} and iDAFormer baselines~\cite{hoyer2022daformer}.

As shown in Table~\ref{table:comparison_mixup} (a) and Table~\ref{table:comparison_mixup} (b), with different basic models, \emph{i.e.,} DeepLabV2~\cite{chen2018deeplab} and SegFormer~\cite{xie2021segformer}, the results demonstrate the superiority of our CAMix to different domain mixup methods.
The main reasons lie in the following aspects:
firstly, we analyze that using CowMix~\cite{french2020milking} results in the occurrence of partial objects in the mixed images, which are hard to learn in the training process. Secondly, CutMix~\cite{french2019semi}, DACS~\cite{tranheden2020dacs} and DAFormer~\cite{hoyer2022daformer} tend to result in severe label contamination and category confusion when generating the mixed results, thus leading to negative transfer.
Besides, iDACS~\cite{tranheden2020dacs} and iDAFormer~\cite{hoyer2022daformer} lack sufficient supervision and produce error-prone target pseudo labels, leading to less-desired performances. In contrast, we exploit the context-dependency across domains as important prior knowledge for facilitating the adaptability toward the target domain, which is largely overlooked by prior works.

\subsection{Ablation Studies and Analysis}
\label{sec:ablation}
In this section, we study the effectiveness of each component in our approach and investigate how they contribute to the final performance from GTAV~\cite{stephan2016gtav} to Cityscapes~\cite{cordts2016cityscapes}.

\noindent \textbf{Effectiveness of CMG:} 
CMG is a  fundamental component of our framework, which is designed to capture the shared context-dependency across domains for CAMix. \textit{Spatial prior (SP)} and \textit{contextual relationship (CR)} are two key components of CMG. The ablation studies of each component in CAMix are reported in Table~\ref{table:ablation_component_level} (a).  Compared to iDACS~\cite{tranheden2020dacs} baseline that performs domain mixup in the target-to-source direction, SP and CR could successfully bring $1.6\% $ and $1.4\% $ of improvements, achieving $53.1\%$ and $54.5\%$ on the former two levels, respectively. By adding the SRC loss on the SigMask level, we can achieve an even higher performance of $55.2\%$.

\noindent \textbf{Effectiveness of different levels:}
Table~\ref{table:ablation_component_level} (b) lists the impacts of different levels on the two settings, \emph{i.e.,} taking GTAV~\cite{stephan2016gtav} and SYNTHIA~\cite{ros2016synthia} as source domains, respectively. Mean Teacher (MT)~\cite{Mean_teacher} baseline achieves $43.1\%$ and $45.9\%$ on two benchmarks, respectively. In-Out means using both the input and output level mixup. By performing CAMix in the input and output level, our method respectively brings $+11.4\%$ and $+13.1\%$ improvements, reaching $54.5\%$ and $59.0\%$. 
By integrating CAMix on three levels together, we finally achieve $55.2\%$ and $59.7\%$ mIoU, respectively. It also reveals that domain mixup in different levels are complementary and together they promote the performance.

\begin{table}[t]
\caption{Ablation study of the SRC loss.}
\label{table:ablation_src}
\begin{center}
\resizebox{0.45\textwidth}{!}{%
\begin{tabular}{c|c|c} \toprule
{$\mathcal{L}_{con}$}  & mIoU & $\triangle$ \\
\midrule
Ours (w SRC loss) & 55.2 & -\\
Ours (w/o SRC loss + MSE Loss) & 44.5 &$\downarrow$ 9.7  \\
 Ours (w/o SRC loss + CE Loss) & 54.2 &$\downarrow$ 1.0\\
\bottomrule
\end{tabular}
}
\end{center}
\end{table}

\begin{figure}[t]
\centering
\includegraphics[scale=1]{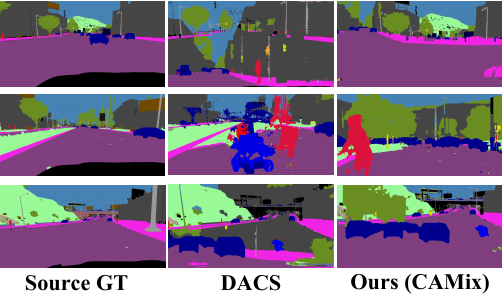}
\caption{
Visual comparisons of different domain mixup algorithms. We use the same source image and target image in each row. \textbf{Left}: source ground truth. \textbf{Middle}: mixed results of DACS~\cite{tranheden2020dacs}. It cannot place the semantic categories in an appropriate context, which results in label contamination and category confusion. \textbf{Right}: Mixed results of our proposed method (CAMix), which can effectively mitigate these issues. }
\label{fig:visulization_mixup}
\vspace{-5mm}
\end{figure}

\noindent \textbf{Effectiveness of SRC:} Table~\ref{table:ablation_src} shows the contribution of the SRC loss on the GTAV $\rightarrow$ Cityscapes benchmark. The full CAMix with all three levels and SRC loss achieves $55.2\%$. If we directly replace the SRC loss with a normal \textit{mean square error (MSE)}, the result is even worse and only reaches $44.5\%$. Using the \textit{cross-entropy (CE)} as the consistency loss boosts the mIoU to $54.2\%$, which is still $1.0\%$ worse than our SRC loss in Eq.~\eqref{eq:8}. The main benefits of the SRC loss are reflected as follows. The SigMask-level domain mixup with the SRC loss could further decrease the uncertainty of the teacher model and promote the teacher model to transfer reasonable knowledge to the student, thus improving the performance. As such, our approach tends to be more stable and effectively ease these negative impacts, \emph{i.e.,} training instability and early performance degradation, during the adaptation process.

\begin{table}[t]
\caption{
Ablation study of different meta class lists. }

\label{table:ablation_meta_class}
\begin{center}
\resizebox{0.50\textwidth}{!}{%
\begin{tabular}{ccccc|c} \toprule
Group I & Group II & Group III & Group IV &  Group V & mIoU \\
\midrule
$\surd$  &  &  & & &68.8 \\
$\surd$  & $\surd$ & &  & & 70.0\\
$\surd$  & $\surd$ & $\surd$ & &  & 69.1\\
$\surd$  & $\surd$  & $\surd$ & $\surd$ & & 67.8 \\
$\surd$  & $\surd$  & $\surd$ & $\surd$ & $\surd$ & 67.3 \\
\bottomrule
\end{tabular}
}
\end{center}
\vspace{-4mm}
\end{table}

\begin{figure}[t]
\centering
\includegraphics[scale=0.5]{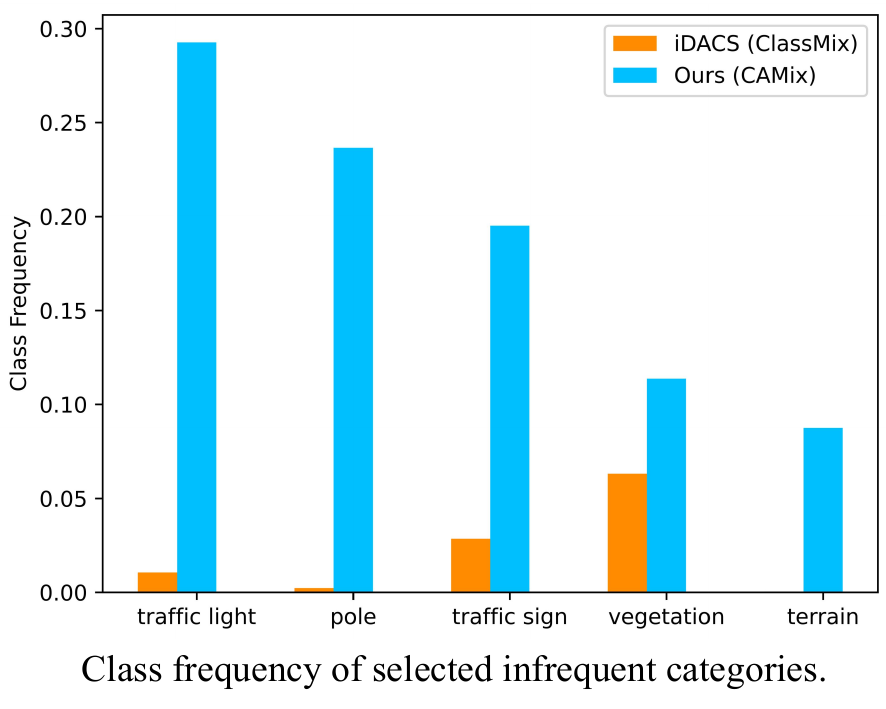}
\vspace{-2mm}
\caption{
Visualizations on class frequency of selected infrequent categories during the mixup in GTAV~\cite{stephan2016gtav} $\rightarrow$ Cityscapes~\cite{cordts2016cityscapes}. 
}

\label{fig:visulization_frequency}
\end{figure}

\noindent \textbf{Ablation of different meta class lists:}
As mentioned in Section~\ref{sec:cmg}, the meta-class list $m$ involves several groups of meta-classes (Table~\ref{table:meta_class}) chosen from the prior knowledge of the hierarchical contexts given in Cityscapes~\cite{cordts2016cityscapes}. Table~\ref{table:ablation_meta_class} reveals the effect of different combinations of meta class groups in GTAV~\cite{stephan2016gtav} $\rightarrow$ Cityscapes~\cite{cordts2016cityscapes} with DAFormer~\cite{hoyer2022daformer}. From the table, we find that when the number of meta class groups is too small or too large, the performance is
less desired, and we observe that the best performance occurs when choosing the first two groups. The main reasons behind this phenomenon can be explained as follows. Too little context knowledge is not enough to provide sufficient supervision signals to facilitate the adaptation, while too much prior knowledge of context-dependency limits the performance of neural networks due to the fact that too many constraints can easily make the learning fall into local optima. Thus, we set the first two groups as our meta class list $m$ in all experiments to show the robustness of our method.

\begin{figure*}[t]
\centering
\includegraphics[scale=1]{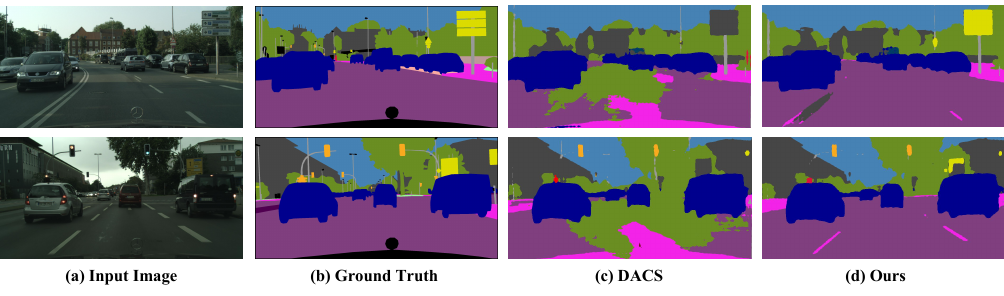}
\caption{Qualitative segmentation results in the SYNTHIA $\rightarrow$ Cityscapes setup. The four columns plot (a) RGB
input image, (b) ground-truth, (c) the predictions of DACS~\cite{tranheden2020dacs} and (d) the predictions of our CAMix. (Best viewed in color.)
}
\vspace{-3mm}
\label{fig:visualization_segmentation}
\end{figure*}

\subsection{Visualization}
\label{sec:visualization}
\noindent \textbf{Visual comparisons of different domain mixup algorithms.} 
As shown in Fig.~\ref{fig:visulization_mixup}, we visualize the mixed samples of different domain mixup algorithms. We use the same source image and target image for each row. We can find that mixed results of previous domain mixup method, DACS~\cite{tranheden2020dacs}, involves some label contamination and category confusion. The main reason is that they overlook the shared context-dependency across domains, and a direct mixup will place the semantic categories in an inappropriate context. Instead, our method (CAMix) explicitly respects the contextual structure of the scenes and generates fewer artifacts in the mixed results. 

\noindent \textbf{Comparisons of class frequency with iDACS~\cite{tranheden2020dacs}}
Fig.~\ref{fig:visulization_frequency} plots the visualizations of class frequency of selected infrequent categories in GTAV~\cite{stephan2016gtav} $\rightarrow$ Cityscapes~\cite{cordts2016cityscapes}.
From this figure, we can observe that: 1)
the frequency of minority classes has significant improvement compared to the iDACS~\cite{tranheden2020dacs} baseline model, leading to improvements in per-class IoU, shown in Table~\ref{table:gtav} and Table~\ref{table:synthia}. 2) 
By considering the contextual relationships, the proposed CAMix strategy can provide sufficient training data and alleviate the over-fitting problem. The main reason behind this phenomenon lies in  the following aspects:
the predictive likelihood of these selected infrequent categories of the iDACS~\cite{tranheden2020dacs} baseline model is low, and these categories are challenging for domain mixup and usually lead to predictions with high uncertainties. With our CAMix, the predictions of these imbalanced categories become confident due to sufficient samples for training.

\begin{figure}[t]
\centering
\includegraphics[scale=0.3]{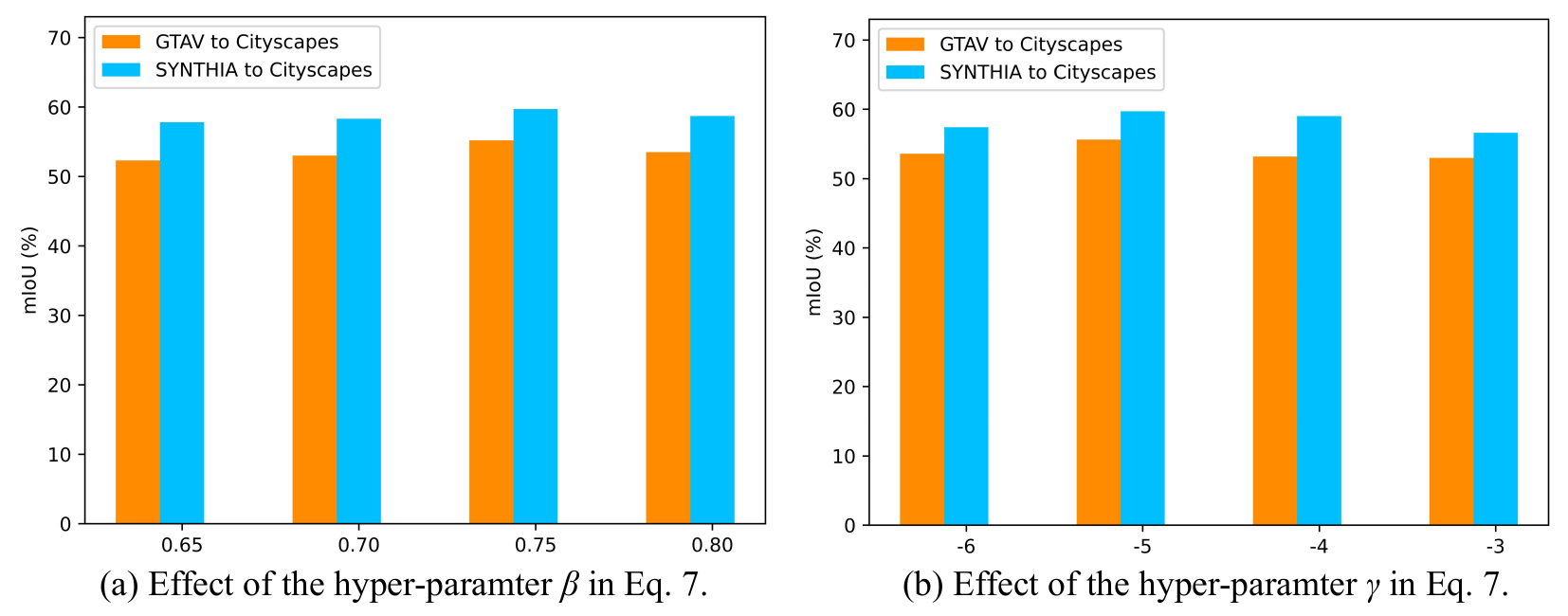}
\caption{
Hyper-parameter analysis of $\beta$ and $\gamma$ in Eq.~\ref{eq:6}.}

\label{fig:visulization_parameter}
\vspace{-5mm}
\end{figure}

\noindent \textbf{Qualitative segmentation results.}
Fig.~\ref{fig:visualization_segmentation} visualizes some segmentation results in the SYNTHIA $\rightarrow$ Cityscapes (16 classes) set-up. The four columns plot (a) RGB input images, (b) ground truth, (c) DACS baseline outputs~\cite{tranheden2020dacs} and (d) the predictions of CAMix.  As we can see from the figure, due to the lack of context-dependency, DACS~\cite{tranheden2020dacs} tends to produce noisy segmentation predictions on some large categories, \emph{e.g.,} ‘road’, ‘sidewalk’, 'truck', etc, and incorrectly classifies some large categories, \emph{e.g.,} the road as sidewalk or terrain, and produces some false predictions on some sophisticated classes, \emph{e.g.,} traffic sign. With the help of our proposed CAMix and SRC loss,  our model manages to produce correct predictions at a high level of confidence. Fig.~\ref{fig:visualization_segmentation} shows that CAMix achieves good performance on ‘road’, ‘sidewalk’, ‘bus’, ‘car’, 'truck', ‘motorcycle’, ‘bicycle’, ‘building’, and ‘terrain’ classes. Our proposed method is capable of outputting high confidence predictions compared to the previous work.

\begin{figure}[t]
\centering
\includegraphics[scale=0.9]{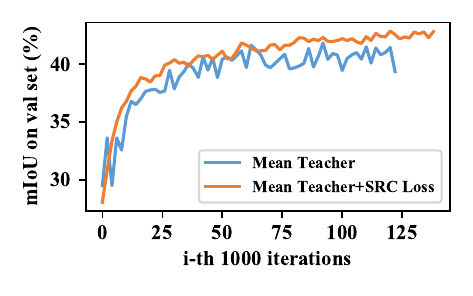}
\vspace{-5mm}
\caption{Performance curve on GTA5~\cite{stephan2016gtav} to Cityscapes~\cite{cordts2016cityscapes}. The blue line corresponds to the conventional consistency regularization~\cite{choi2019self}. The orange line indicates the consistency-based adaptation with our SRC loss. Our method eases the issue of training instability and early performance drop. }
\label{fig:miou_performance_curve}
\vspace{-5mm}
\end{figure}

\noindent \textbf{Analysis of hyper-parameter $\beta$ and $\gamma$ in Eq.~\ref{eq:6}.}
Fig.~\ref{fig:visulization_parameter} plots the performance of models trained with different hyper-parameter  ($\beta$ and $\gamma$) values on the setting of GTAV~\cite{stephan2016gtav} $\rightarrow$ Cityscapes~\cite{cordts2016cityscapes} and SYNTHIA~\cite{ros2016synthia} $\rightarrow$ Cityscapes~\cite{cordts2016cityscapes}. As mentioned above, $\beta$ is the initial state of the dynamic threshold $H$, and $\gamma$ controls the exponential speed of the dynamic threshold.
The highest mIoU on the target domain is
achieved when the value of  $\beta$  is around $0.75$ and $\gamma$ is around $-5$, which means that this initial state and  exponential speed benefit domain adaptation the most. Thus, we simply set the same $\beta=0.75$ and $\gamma=-5$ in all experiments to show the robustness of our method in different settings.

\noindent \textbf{Performance curve of adaptation.}
Fig.~\ref{fig:miou_performance_curve} plots the performance curves to show the effectiveness of SRC loss when adapting from GTAV~\cite{stephan2016gtav} to Cityscapes \cite{cordts2016cityscapes} with VGG16~\cite{simonyan2015vgg} backbone. 
We observe that the curve of Mean Teacher~\cite{choi2019self}, which is representative of previous consistency regularization methods, fluctuates wildly and causes the training instability and early performance degradation. The main reason is that they largely neglect the context knowledge shared by different domains and perform a rough distribution matching, resulting in less-desired performances. Instead, we effectively ease these negative impacts and decrease the uncertainty of the segmentation model, by introducing the SRC loss.

\section{Conclusion}
In this paper, we proposed a novel context-aware domain mixup (CAMix) framework via explicitly exploiting context-dependency across domains to enhance the adaptability for domain adaptive semantic segmentation. We present a contextual mask generation (CMG) strategy, which is critical for guiding the whole pipeline on three different levels, \emph{i.e.,} input level, output level and, significance mask level. Our approach can explicitly explore and transfer the shared context-dependency across domains, thus narrowing down the domain gap. We also introduce a significance-reweighted consistency loss (SRC) to penalize the inconsistency between the mixed student prediction and the mixed teacher prediction, which effectively eases the adverse impacts of the adaptation, \emph{e.g.,} training instability and early performance degradation. Extensive experiments with analysis 
demonstrate that our approach soundly outperforms the state-of-the-art methods in domain adaptive semantic segmentation.


\ifCLASSOPTIONcaptionsoff
  \newpage
\fi

\bibliographystyle{IEEEtran}
\bibliography{egbib}

%


\vspace{-1cm}
\begin{IEEEbiography}[{\includegraphics[width=1in,height=1.25in,clip,keepaspectratio]{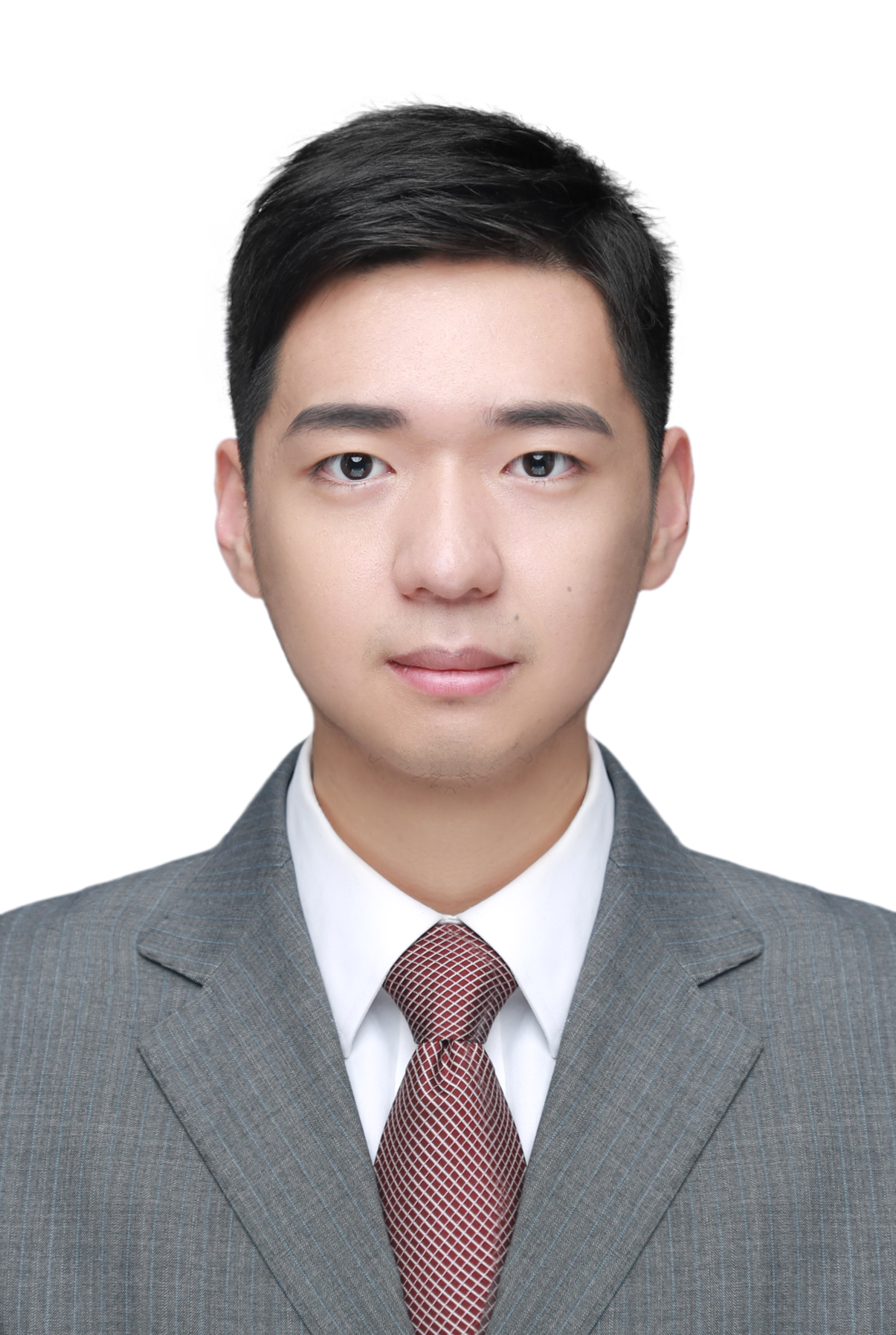}}]{Qianyu Zhou} is currently pursuing his Ph.D. degree in the Department of Computer Science and Engineering, Shanghai Jiao Tong University. Before that, he received a B.Sc. degree in Jilin University in 2019. His current research interests focus on computer vision, scene understanding, domain adaptation. He serves as the reviewer of IEEE TPAMI, IEEE TIP, CVPR, ECCV, AAAI, etc.
\end{IEEEbiography}

\vspace{-1.5cm}

\begin{IEEEbiography}[{\includegraphics[width=1in,height=1.25in,clip,keepaspectratio]{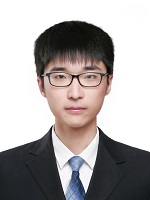}}]{Zhengyang Feng}
is currently pursuing his M.Sc. degree in the Department of Computer Science and Engineering, Shanghai Jiao Tong University. Before that, he received a B.E. degree in information security from Harbin Institute of Technology, Weihai, China, in 2020. His current research interests focus on pattern recognition with limited human supervision.
\end{IEEEbiography}

\vspace{-1cm}
\begin{IEEEbiography}[{\includegraphics[width=1in,height=1.25in,clip,keepaspectratio]{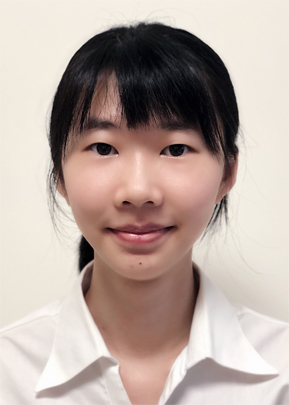}}]{Qiqi Gu}
received a MA.Eng. degree in Department of Computer Science and Engineering, Shanghai Jiao Tong University, in 2022. Her current research interests focus on domain adaptation of object detection and semantic segmentation.
\end{IEEEbiography}

\vspace{-1cm}

\begin{IEEEbiography}[{\includegraphics[width=1in,height=1.25in,clip,keepaspectratio]{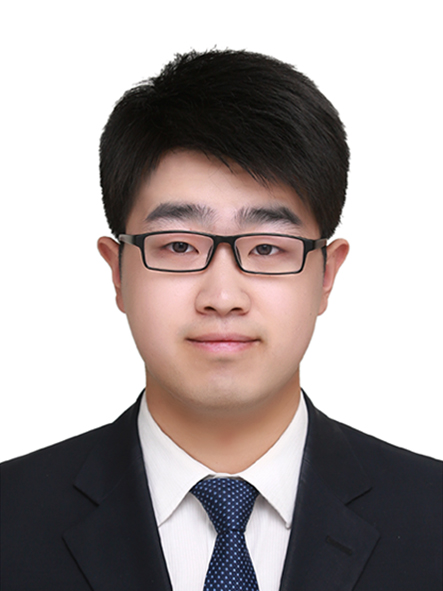}}]{Jiangmiao Pang} is currently a research scientist at Shanghai AI Laboratory. He obtained his Ph.D. degree from Zhejiang University in 2021, and did his postdoc at MMLab, The Chinese University of Hongkong, afterwards. His research interests include computer vision and robotics, especially their applications in autonomous driving.
\end{IEEEbiography}

\vspace{-1cm}

\begin{IEEEbiography}[{\includegraphics[width=1in,height=1.25in,clip,keepaspectratio]{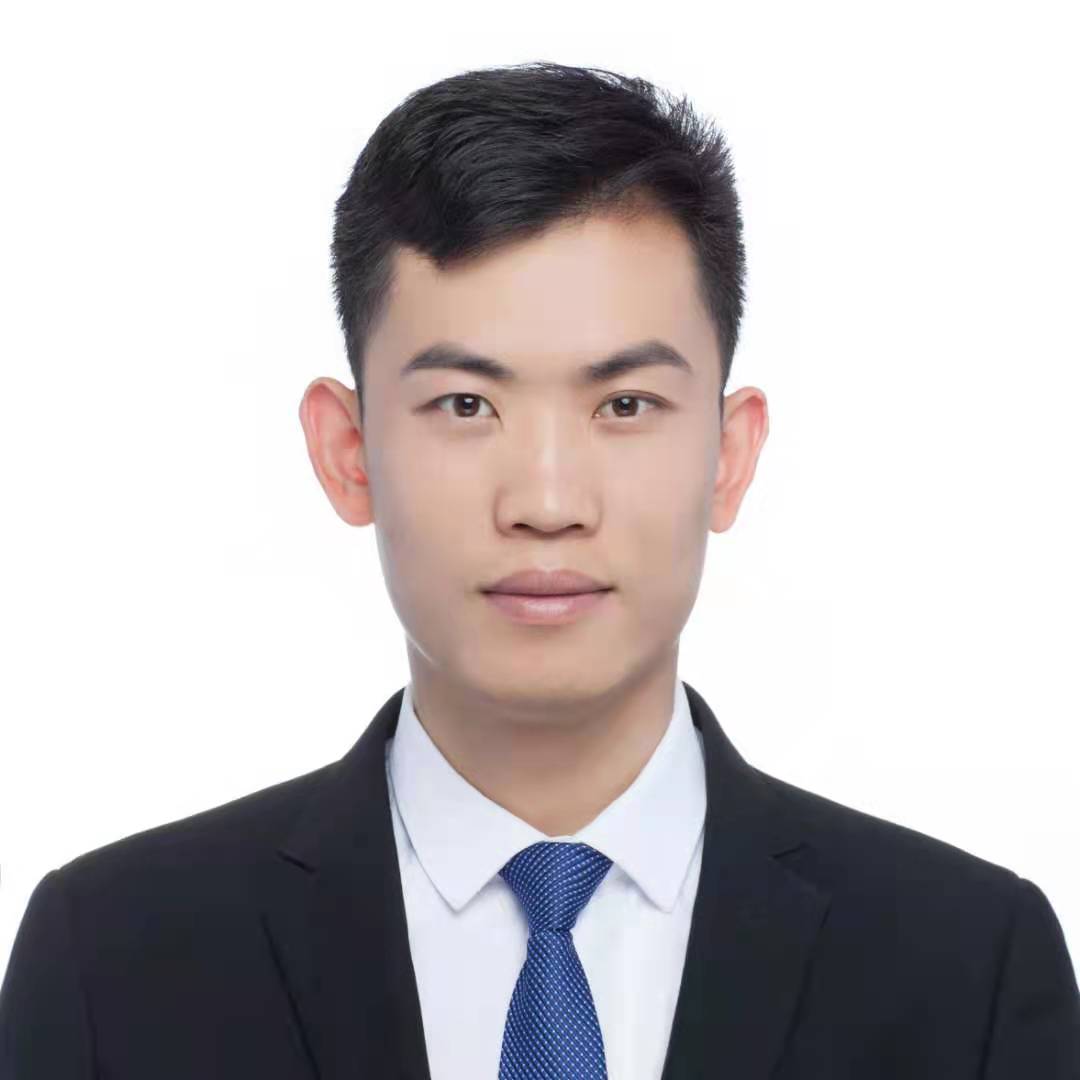}}]{Guangliang Cheng}
is currently a Senior Research Manager in SenseTime. Before that, he was a Postdoc researcher in the Institute of Remote Sensing and Digital Earth, Chinese Academy of Sciences, China, and he received his Ph.D. degree with national laboratory of pattern recognition (NLPR) from the Institute of Automation, Chinese Academy of Sciences, Beijing. His research interests include autonomous driving, scene understanding, domain adaptation and remote sensing image processing.
\end{IEEEbiography}

\vspace{-1cm}

\begin{IEEEbiography}[{\includegraphics[width=1in,height=1.25in,clip,keepaspectratio]{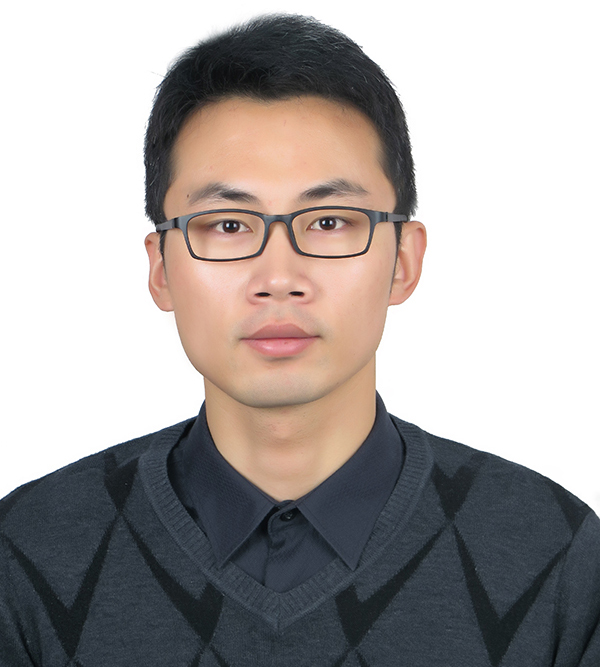}}]{Xuequan Lu}
is an Assistant Professor at the School of Information Technology, Deakin University, Australia. He spent more than two years as a Research Fellow in Singapore. Prior to that, he earned his Ph.D at Zhejiang University (China) in June 2016. His research interests mainly fall into the category of visual computing, for example, geometry modeling, processing and analysis, animation/simulation, 2D data processing and analysis. More information can be found at http://www.xuequanlu.com.
\end{IEEEbiography}

\vspace{-1cm}

\begin{IEEEbiography}[{\includegraphics[width=1in,height=1.25in,clip,keepaspectratio]{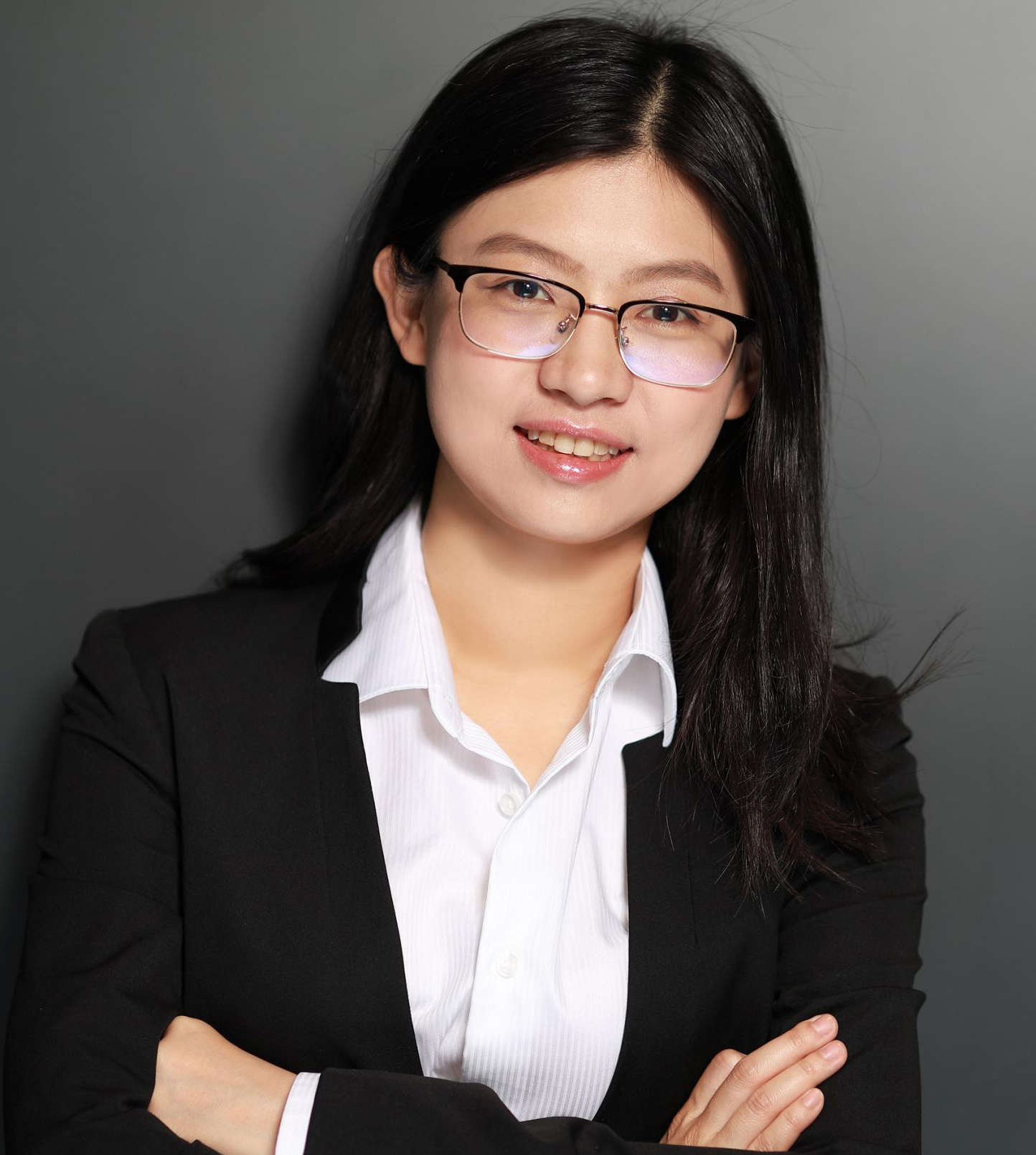}}]{Jianping Shi}
is an Executive Research Director at SenseTime. Currently her team works on developing algorithms for autonomous driving, scene understanding, remote sensing, etc. She got her Ph.D. degree in Computer Science and Engineering Department in the Chinese University of Hong Kong in 2015 under the supervision of Prof. Jiaya Jia. Before that, she received the B. Eng degree from Zhejiang University in 2011. She has served regularly on the organization committees of numerous conferences, such as Area Chair of CVPR, ICCV, etc.
\end{IEEEbiography}

\vspace{-1cm}

\begin{IEEEbiography}[{\includegraphics[width=1in,height=1.25in,clip,keepaspectratio]{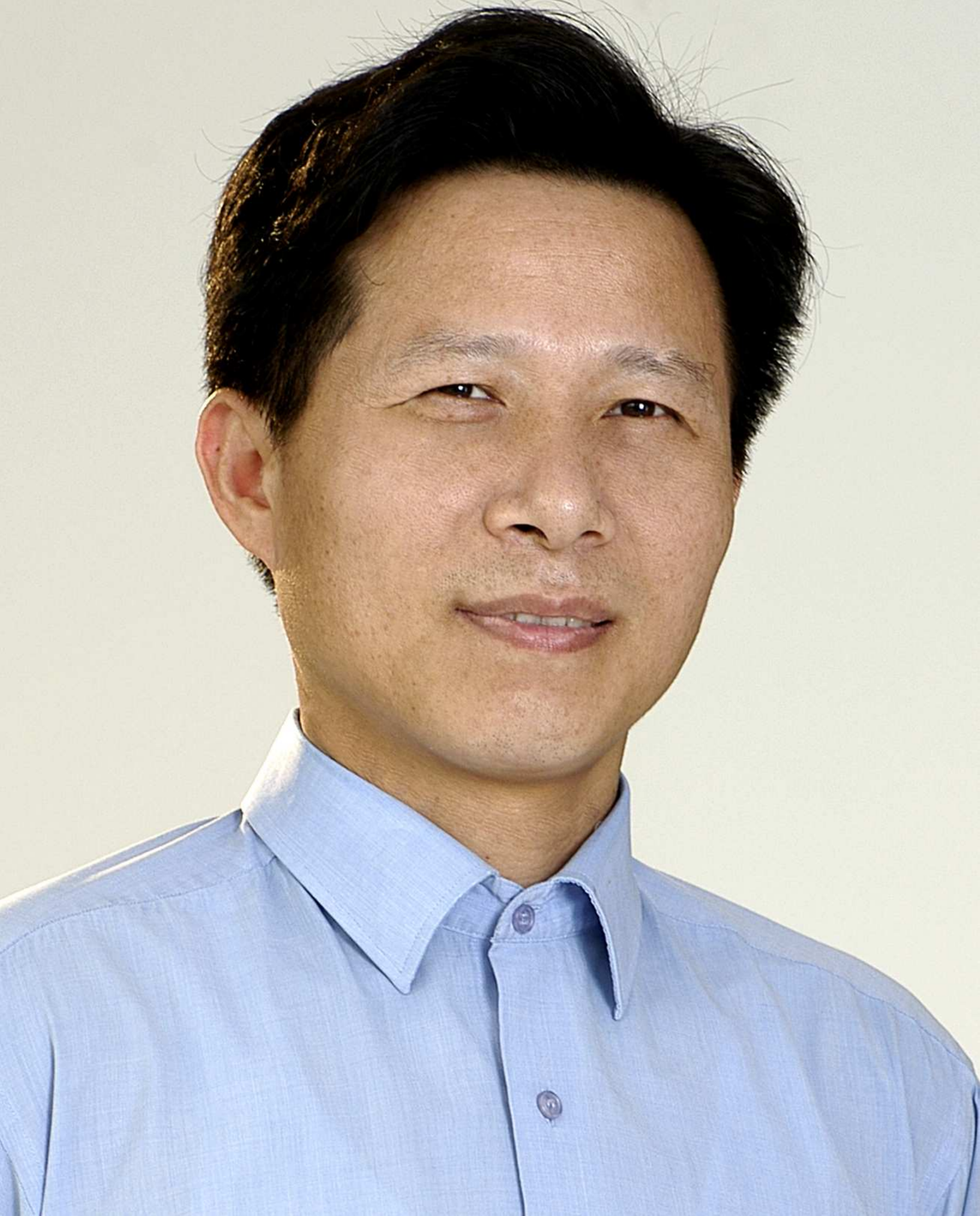}}]{Lizhuang Ma} is now a Distinguished Professor, Ph.D. Tutor, and the Head of the Digital Media and Computer Vision Laboratory at the Department of Computer Science
and Engineering, Shanghai Jiao Tong University, China. He received his B.S. and Ph.D. degrees from the Zhejiang University, China in 1985 and 1991, respectively.
He was also a Visiting Professor at the
Frounhofer IGD, Darmstadt, Germany in 1998,
and was a Visiting Professor at the Center for
Advanced Media Technology, Nanyang Technological University, Singapore from 1999 to 2000. He has published more than 200 academic research papers in both domestic and international
journals. His research interests include computer aided geometric design, computer graphics, computer vision, scientific data visualization, computer animation, digital media technology, and theory and applications for computer graphics, CAD/CAM.  
\end{IEEEbiography}

\end{document}